\documentclass[preprint,12pt,authoryear]{elsarticle}

\usepackage{amssymb}

\journal{Entertainment Computing}

\begin{document}

\begin{frontmatter}



\title{Dynamic Difficulty Adjustment on MOBA Games}


\author{Mirna Paula Silva, Victor do Nascimento Silva and Luiz Chaimowicz}

\address{Department of Computer Science\\
Universidade Federal de Minas Gerais (UFMG)\\
Belo Horizonte, Brazil}




\begin{abstract}
This paper addresses the dynamic difficulty adjustment on MOBA games as a way to improve the player’s entertainment. Although MOBA is currently one of the most  played genres around the world, it is known as a game that offer less autonomy, more challenges and consequently more frustration. Due to these characteristics, the use of a mechanism that performs the difficulty balance dynamically seems to be an interesting alternative to minimize and/or avoid that players experience such frustrations. In this sense, this paper presents a dynamic difficulty adjustment mechanism for MOBA games. The main idea is to create a computer controlled opponent that adapts dynamically to the player performance, trying to offer to the player a better game experience. This is done by evaluating the performance of the player using a metric based on some game features and switching the difficulty of the opponent's artificial intelligence behavior accordingly. Quantitative and qualitative experiments were performed and the results showed that the system is capable of adapting dynamically to the opponent's skills. In spite of that, the qualitative experiments with users showed that the player's expertise has a greater influence on the perception of the difficulty level and dynamic adaptation.


\end{abstract}


\end{frontmatter}


\section{Introduction}

The game industry is growing at a fast pace, globally generating more revenue than film and music industries \citep{thompson2015interrogating}. Games are considered a great source of entertainment \citep{nareyeka} and, due to that, the industry is increasingly investing more resources in research and development. This allows developers to create realistic graphics, deep narratives and complex artificial intelligence (AI), leading to games even closer to reality \citep{machadoplayerA, smithtaxonomy}. 

The development of realistic games results in an improved player immersion which, in general, increases their satisfaction~\citep{bowman2007virtual}. Although this is a well explored approach, it is not the only way to make games more attractive. According to \cite{Yannakakis07towardsoptimizing}, the player's psychological factor makes direct influence to this attractiveness, requiring the game to maintain the player interested on it. An approach to captivate the player into the game experience is to make the challenges directly associated to the player's skill \citep{feijo2013evaluating}. However, a game may not suit the expectation of players with different skills. While a player may have a hard time in final levels of a game, there may be another player that cannot win the initial ones. This scenario requires that the game dynamically adjusts itself presenting challenges that suit the needs and skills of each player. This game adjustment can be performed by a technique called dynamic difficulty adjustment (DDA) or dynamic difficulty balancing.

In spite of different studies in DDA \citep{stanley2005evolving, spronck2006adaptive, togelius2007towards,bakkes2009rapid, wheat2015dynamic}, none of them tackles MOBA (Multiplayer Online Battle Arena) games, which are one of the most played games genres nowadays, having almost 30\% of online computing gameplay time\footnote[1]{http://goo.gl/zgKjJL}. Although they are very popular among gamers, there is not much attention from researchers over this game genre. This can be related to the inherent challenges of developing competitive artificial intelligent agents for MOBA games as well as the constant updates and changes on these games.

This paper presents a dynamic difficulty adjustment mechanism for MOBA games. The main idea is to create a computer controlled opponent that adapts dynamically to the player performance, trying to offer to the player a better game experience. This is done by evaluating the performance of the player using a metric based on some game features and switching the difficulty of the opponent's artificial intelligence behavior accordingly. This idea was initially proposed in \citep{silva2015b, silva2015a}, and here we revisit the mechanism, giving more details about its implementation and performing a set of experiments with human players in order to have a qualitative evaluation. We also present and discuss the main characteristics and challenges of MOBA games, trying to encourage other researchers to use them as testbeds in their future work. 

This paper is organized as follows: in Section 2 we present the related work and background on difficulty balance; Section 3 covers MOBA games aspects, history and challenges, as well as the game DotA used as testbed in this work; Section 4 addresses the methodology and the proposed mechanism; Section 5 discusses the performed experiments of agents versus agents and the obtained results; Section 6 presents the experiments performed with users and the collected results; and finally, Section 7 brings the conclusion and directions for future work.
\section{Difficulty Balance}

Difficulty balance, or difficulty adjustment, consists on doing modifications to parameters, scenarios and/or game behaviors in order to avoid the player's frustration when facing the game challenges \citep{feijo2013evaluating, koster2010theory}. According to \cite{mateas2002interactive} and \cite{hunicke2005case}, it is possible to adjust all game features using the correct algorithms, from storytelling to maps and level layouts, all online. These adjustments allow the game to adapt itself to each player, making he/she entertained throughout the game. To make this possible, \cite{andrade2005extending} describes that the dynamic difficulty adjustment must attend three basic requirements. First of all, the game must automatically identify the players' skills and adapt to it as fast as possible. Second, the game must track the player's improvement and regressions, as the game must keep balance according to the player's skill. At last, the adaptive process must not be explicitly perceived by players, keeping game states coherent to previous ones. However, before applying the dynamic difficulty adjustment, it is necessary to understand the meaning of difficulty.


The meaning of difficulty is abstract in many ways and some aspects should be taken into account to evaluate and measure difficulty. For this measuring, we can consider level design characteristics \citep{bartle2004designing}, amount of resource or enemies \citep{hunicke2005case}, amount of victories or losses \citep{poole2004trigger, xavier2010condiccao}, among other metrics. Nevertheless, dynamic difficulty adjustment is not as simple as just giving player additional health items when in trouble. This problem requires estimation of time and intervention in the right moment, since maintaining the player entertained is a complex task in an interactive context \citep{hunicke2005case}.

A wide range of tasks and challenge levels can be found in games. For example, tasks that require high skill and synchronism (First Person Games), tasks that require logic and problem solving skills (Puzzles), tasks related to planning (Strategy games), and so on \citep{klimmt2009player}. According to \cite{klimmt2009player}, there is evidence that the completion of tasks and challenge overcoming are directly related to player satisfaction and fun. \cite{yannakakis2008model} developed a study about the most popular approaches for player modeling during interaction with entertainment systems. According to this study, most qualitative approaches proposed for player entertainment modeling tends to be based in conceptual definitions proposed by \cite{malone1981toward} and \cite{csikszent1991flow}.

\cite{malone1981toward} defended the need for a specific motivation during gameplay to entertain the player. The necessary features to reach such motivation are: {\em fantasy, control, challenges and curiosity}. The use of {\em fantasy} as part of game world could improve player motivation, creating objects, scenarios or situations that the player could explore. {\em Control} is a player feeling through which he/she is part of game control. Given the interaction of games, all of them make the player feel involved in game control and the control levels can change from game to game. {\em Challenge} implies that the game should pursue tasks and goals in an adequate level, making the player feel challenged to his/her limits. The uncertainty of completing tasks or goals provided by game mechanics encourages the player motivation. Finally, {\em curiosity} suggests that game information must be complex and unknown, to encourage exploration and reorganization of information by players. Games must pursue multiple situations or scenarios from the main course since it helps to stimulate the player to explore the unknown \citep{malone1981toward, egenfeldt2013understanding}.

\begin{figure}[ht]
	\centering
		\includegraphics[height=5.5cm]{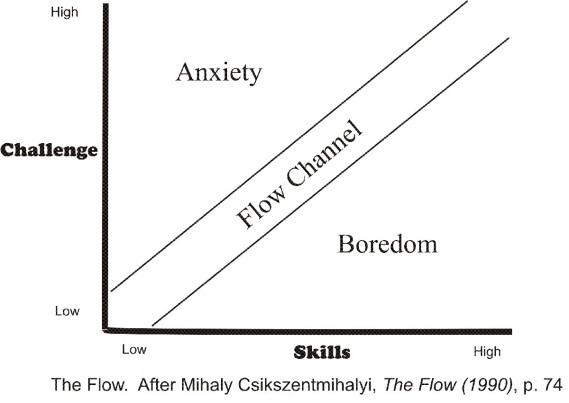}
	\caption{Diagram of flow theory, by Csikszentmihalyi.}
	\label{fig:flow}
\end{figure}

The qualitative approach proposed by \cite{csikszent1991flow} is called {\em flow theory} or {\em flow model}. According to the author, {\em flow} is a mental state experienced when the user is executing an activity in which he/she is immersed, feeling focused, completely involved and fulfilled during task execution. So, this model takes into account the psychological steps that players reach during gameplay. In this sense, the main goal is controlling the challenge levels aiming to maintain the player inside the flow, avoiding to reach boredom (no challenges at all) or frustration (challenges are too hard). Figure \ref{fig:flow} show a graph of flow theory presented by \cite{csikszent1991flow}.

The model presented by Csikszentmihalyi shows how a task difficulty is directly related to the perception of who is executing it. The flow channel illustrates that difficulty can be progressively improved, since there exists time to the player to learn and improve his/her skills to overcome this challenge \citep{csikszentmihalyi2000beyond}. Thereby, this model avoids frustration of very hard situations or boredom caused by very easy situations. Furthermore, \cite{csikszentmihalyi2010effortless} go beyond and determine that the ratio of challenges to skills should be around 50/50 in order to produce enjoyable experiences. 

On the other hand, there are some studies that question if the ratio of challenges to skills is really a measure of flow. \cite{lovoll2014can}, for instance, present a work with some empirical evidence that contests the idea that flow is produced when challenges and skills are harmonized. According to them, the interaction between challenges and skills as independent variables 
does not support the challenge skill ratio proposed by Csikszentmihalyi and Nakamura.

In a different approach, if we can balance the fantasy, control, challenge and curiosity proposed by \cite{malone1981toward} and associate it to the progressive development of difficulty presented by \cite{csikszent1991flow}, it is possible that the resulting game can entertain the player. However, using just these features is not sufficient to show if game challenges are compatible with player skills. So, it is necessary measuring techniques to define when and how difficulty should be adjusted.
\subsection{Evaluating the Difficulty Level}

According to \cite{andrade2005extending}, there are some different approaches to dynamically balance the difficulty level of a game. However, all of these approaches require measuring, implicitly or explicitly, the difficulty level that the player is facing on that moment. 
This measurement can be done by using heuristics, for example the success rate of skill landing, the capture of enemy points, the time used to complete a task or any other metric that can evaluate the player. \cite{missura2009player} made a relation between game runtime, health and score in a way that it composes an evaluation criteria that performs the game difficulty adjustment. \cite{demasi2003line} developed a heuristic function called ``Challenge Function" that is responsible for describing the game state, and tries to show how hard the game is for the player in a given time.

Another way to track difficulty levels is using some physiological signs, informally called {\em body language}. \cite{van2008exploring} mentions that the body language of a player could be related to his/her experience during play. According to the authors, there are evidences that show that specific postures, facial expressions, eye movements, stress over mouse / keyboard / joystick, and others, could evidence experiences like interest, excitement, frustration and boredom. For the evaluation of player experience, authors created a monitoring ambient, placing pressure sensors at different devices (mouse, chair, etc). Also cameras were placed to register movements and facial expression. The results of this experiment show that the behaviors observed are directly related to the excitement level and dominance felt during the game. \cite{nacke2008flow}, besides using cameras to capture body language, also used electrodes to track mental reaction from players during a First Person Shooter (FPS) match. The results obtained during player monitoring were based on the flow theory proposed by \cite{csikszent1991flow}, therefore, authors could observe if the players were inside the flow, anxious or bored during the gameplay.

Although the explicit measuring (external monitoring) of difficulty levels could provide fine results related to game fitness to player's skill, it is impracticable to the dynamic difficulty adjustment. Not all players have measuring tools at home and using such tools could be intrusive, since this could make the player uncomfortable by being monitored. Implicit approaches (metrics and heuristics) do not need external equipment, therefore these approaches are more popular among game developers. Besides, they contribute to the fact that players must not perceive that difficulty is being adjusted during gameplay.

This paper tries to perform a dynamic difficulty adjustment through the development of a mechanism that switches between three distinct levels of artificial intelligence in order to provide an opponent that better suits the player's abilities. The mechanism performs several evaluations during the match, detecting the moments in which the game is unbalanced, and then executes the difficulty adjustment.


\section{Multiplayer Online Battle Arena}

Multiplayer Online Battle Arena, also known as Action Real-time Strategy, or simply as MOBA, is a genre originated from Real-Time Strategy (RTS), as a modification of the original game. The first known MOBA game is {\em Aeon of Strife}, created from the game Starcraft. In this game, the player should choose an unit and work his/her way to conquer the enemy's base with the chosen unit and its special powers. This game structure was maintained through the improvement of MOBAs. One interesting fact is that these games came up from simple fan made games to become one of the most played genres in the world, as will be discussed later in this section. 

Another interesting fact is that according to \cite{johnson2015all} and \cite{kwak2015}, MOBA games were found to offer less autonomy, more frustration and more challenges to players. These findings with respect to autonomy seems most likely to be a function of the fact that MOBA games involve fairly focused competition with other players. Moreover, the greater levels of frustration experienced may also be a function of the focused competition that occurs in MOBA games and the steep learning curve. With less focus on the qualities of the game and greater focus on competing and cooperating with others, there is more potential for frustration with the performance of other players. This interpretation is supported by players reporting a greater challenge when playing MOBA games \citep{nascimento2015b}. Due to these characteristics, the use of a mechanism that performs the difficulty balance dynamically seems to be a viable alternative to minimize and/or avoid that such frustrations be experienced by the players.

In this section we present the MOBA history, its characteristics and gameplay. Then, we discuss the unique features present in DotA, the selected platform to be used as testbed in this work, and why it is so hard to develop AI agents to play against humans on these games.

\subsection{History}

As Real-Time Strategy (RTS) games became popular, we observed the urge of the players to create their own maps and gameplay styles. This phenomena resulted in a community of developers that later came to be known as {\em modders}. Their work was known as {\em mods}, an acronym to the word ``modification''. In those mods, players could use the original game environment to play by their own rules, allowing them to create a fantasy world beyond the limits of the original game. 

Released in 1998, {\em Aeon of Strife} was the first mod to present a unique characteristic that caught RTS players attention. Instead of focusing on resource collection and base construction, the mod valued the player ability to control a single unit, an ability known as {\em micromanagement}. This characteristic invited players to duel against each other into single or teams battles, showing to be a successful approach to get players involved in the game.

\begin{figure}
\centering
\includegraphics[height=8cm]{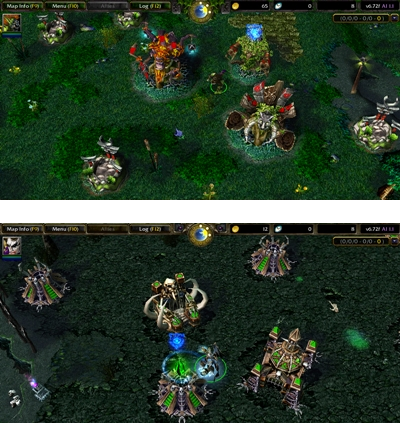}
\caption{The Sentinel base (top) and the Scourge base (bottom).}
\label{fig:bases}
\end{figure}

{\em Aeon of Strife} inspired many other mods that followed its guidelines. Later in 2005, one of those mods stood out in the crowd: {\em Defense of the Ancients} (DotA). The platform was not Starcraft anymore, but another game developed by Blizzard: Warcraft III. The game had the perfect environment and an open API that allowed the modders to do their job. Therefore, they created the DotA map where players would assume the control of a single unit called {\em Hero} and develop this unit by defeating enemies, just like in a Role Playing Game (RPG). Every hero has a single set of attributes and powers, characterizing them in a role. The player could then choose a hero based on its team needs or on its own gameplay style. The game story of the DotA map were also inherited from the Warcraft myth: the war between two races in the Warcraft world, the Night Elf and the Undead. Thereby, players were invoked to defend a main structure called {\em Ancient}, which must be destroyed in order to achieve the game goal. The game is divided into two teams with five players each: The Sentinel, having the Tree of Life as Ancient; and The Scourge, having the Frozen Throne as Ancient. Screen shots of the team bases can be found in Figure \ref{fig:bases}.

The DotA popularity among players resulted in behavioral changes in the general gameplay. Instead of just playing on LANs, players were excited about playing on the Internet. At that time, the broadband was expanding all around the world and players wanted to test it, as well as test their skills challenging others around the world. There were platforms like Garena that had dedicated servers only for DotA matches. DotA's gameplay became so famous that it inspired the game industry to create professional games based on this play style. In 2009, Riot Games released a game called {\em League of Legends} (LoL) \citep{riot2009lol}, with characteristics very similar to DotA. This company created the term MOBA, referring to their debuting title as a Multiplayer Online Battle Arena. Later, Valve has released its own game, known as {\em Dota2}, that immediately caught the attention of the world and media because of the 1 million Dollar tournament. Lastly, around 2010, S2 games have released its own title, {\em Heroes of Newerth}, that has similar characteristics to {\em DotA}, {\em Dota2} and {\em LoL}.
Nowadays, there are other titles, such as {\em Strife} and {\em Heroes of the Storm}, but they did not get many players as the games released before. 

\begin{figure}
\centering
\includegraphics[height=8cm]{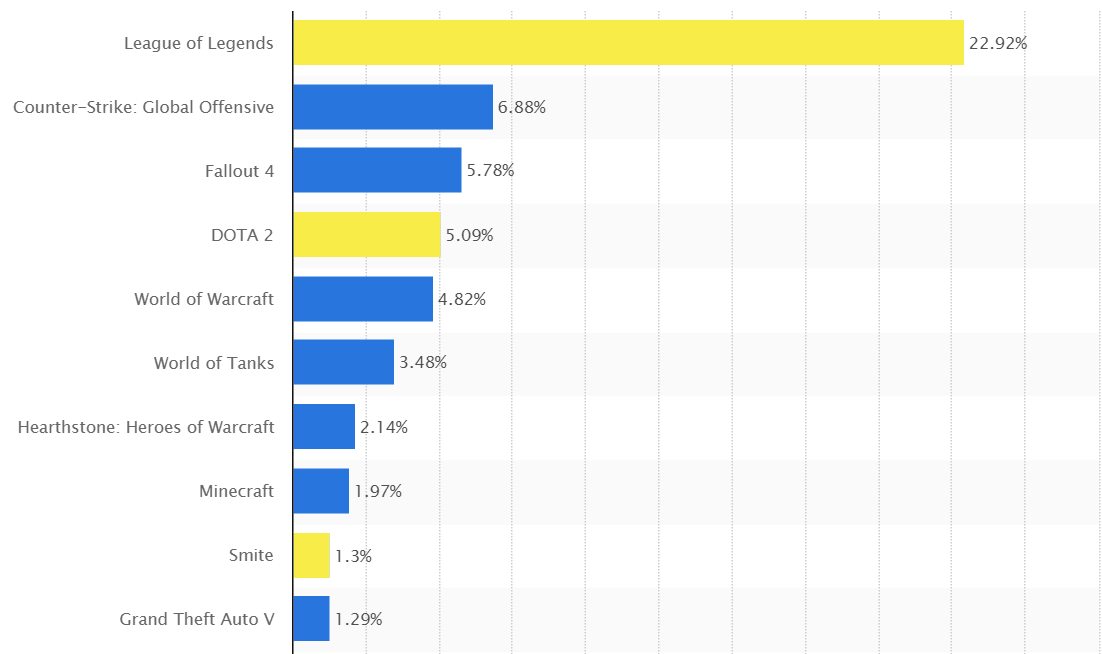}
\caption{Ten most played games of 2015. Source: Raptr/Statista}
\label{fig:mostplayed}
\end{figure}

In numbers, we can see that MOBA genre is a world success, as shown in Figure \ref{fig:mostplayed}. By 2012, the game {\em League of Legends} has overcome {\em World of Warcraft} as the most played game in the world \citep{gaudiosi2012riot}. In November 2015, as reported by Raptr, {\em League of Legends} alone represented more than 22\% of the worldwide gameplay\footnote[1]{http://goo.gl/zgKjJL}. There are international eSports competitions involving those games and millionaire prizes. In 2015, for example, the official Valve's World Tournament of {\em Dota2} called ``The International'' distributed a total of US\$ 18 million\footnote[2]{http://goo.gl/6iXfMD}.

\subsection{Gameplay and Characteristics}

To provide challenges that suit the player's skills it is necessary to comprehend the gameplay that involves the game. The MOBA game can be summarized into two teams playing against each other: Team 1 and Team 2. Players on the Team 1 are based at the southwest corner of the map, and those on the Team 2 are based at the northeast corner. Each base is defended by towers and waves of NPC units (called creeps) that guard the main paths leading to their base, called lanes. In the center of each base there is one main structure. This structure is the goal of the game, which the enemy must destroy in order to win the match.

The teams are composed by five players, where each player controls one specific and powerful unit with unique abilities, which is called Hero or Champion. In most MOBAs, players on each team choose one from dozens of heroes, each with different abilities and tactical advantages over the others. The scenario is highly team-oriented: it is difficult for one player to lead the team to victory by himself/herself. 

Since the gameplay goes around strengthening individual heroes, it does not require focus on resource management and base-building, unlike most traditional RTS games. When killing enemy or neutral units, the player gains experience points and when enough experience is accumulated the player increases his/her level. Leveling up improves the hero's toughness and the damage it inflicts, allowing players to upgrade spells or skills. 

In addition to accumulating experience, players also manage a single resource of gold that can be used to buy items such as armory, potions, among others. Besides a small periodic income, heroes can earn gold by killing hostile units, towers, base structures, and enemy heroes. With gold, players can buy items to strengthen their hero and gain abilities. Also, certain items can be combined with recipes to create more powerful items. Buying items that suit the chosen hero is an important tactical element of the game.

\subsection{Map}

\begin{figure}
\centering
\includegraphics[height=7cm]{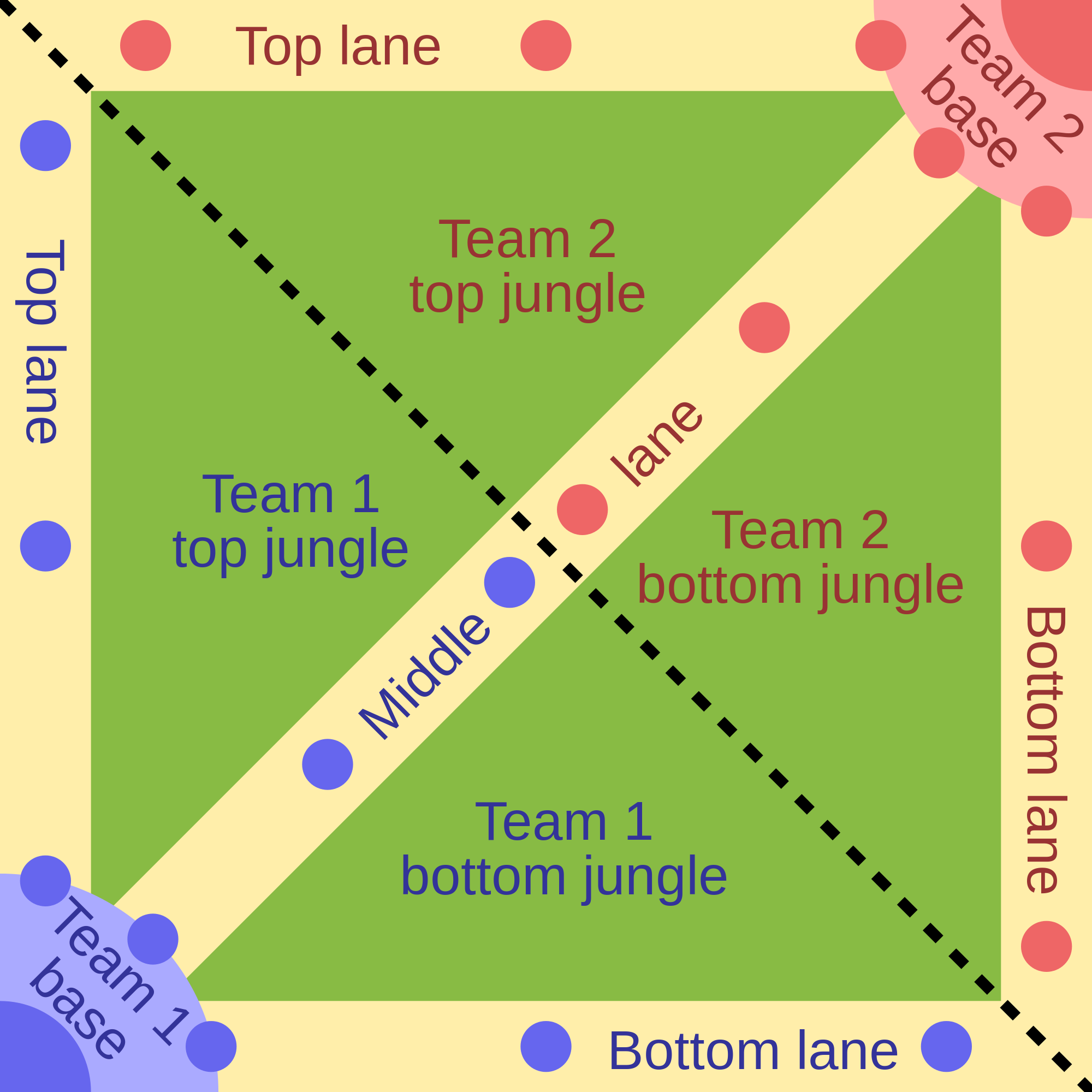}
\caption{General map layout from MOBA games.}
\label{fig:MOBAmap}
\end{figure}

The map is segmented into three different lanes, the top, the bottom, and the middle lane. Each one of these lanes leads to the other team's base, guarded by towers along the way. Figure \ref{fig:MOBAmap} represents a general MOBA map with its lanes, bases and towers along each lane.

The map area located between the lanes is called jungle. This is where neutral creeps can be found, which can be killed for gathering more gold and experience points. It is possible to level up by killing creeps in the jungle instead of in the lanes. This practice is called jungling.

During the early laning phase of the game, most gameplay is centered around ``farming''. It means that players focus on collecting resources and leveling up their heroes by defeating enemy units, like creeps or heroes. In the case of the junglers, they walk through the jungle and kill neutral units. Further, the junglers and their team seek for failures on the enemy teams' strategy, looking for catching them in traps or  performing gang killing, the so called {\em ganks}. Lastly, there is the late game phase, when the gameplay is commonly focused on teamfights, {\em i.e.}, teams use their heroes to fight in groups, looking for weakening the enemy team and pushing the lane towards the enemy's base.

Each team has defensive towers placed along the lanes leading to the Ancient. Those towers inflict heavy single target damage to heroes and creeps. In the early stages of the game, a hero can only take a few hits from a tower before dying, so one must be careful as to not get in a bad positioning relatively to the towers until they have gained enough strength. In the Figure \ref{fig:MOBAmap} the towers are represented by little circles placed in the lanes.

\subsection{\textsc{Defense of the Ancients}}

The game \textsc{Defense of the Ancients} (DotA) is a Multiplayer Online Battle Arena (MOBA) mod version of the game \textsc{Warcraft III: Reign of Chaos} and later to its expansion, \textsc{Warcraft III: The Frozen Throne}. The scenario objective is for each team to destroy the opponents' Ancient, heavily guarded structures located at opposing corners of the map. Players use powerful units known as heroes, and are assisted by allied heroes (played by other users) and AI-controlled fighters known as creeps. As in role-playing games, players level up their heroes and use gold to buy items and equipment during the match.


Moreover, since DotA is a mod of the game \textsc{Warcraft III: Reign of Chaos}, it becomes easier to modify because we can use the tools made to edit \textsc{Warcraft} maps to do it. Therefore, the game \textsc{Defense of the Ancients} (DotA) was chosen to be the testbed of this work.

\subsection{Game Adaptations}

To use the game \textsc{Defense of the Ancients} as a testbed, some adaptations were made in order to better suit the needs of this work. The original game allows the player to choose his/her hero among 110 different options. But, for this work, we chose to restrict this quantity to only 10 heroes, equally distributed between both teams.

\begin{figure}[h]
\centering
\includegraphics[height=6cm]{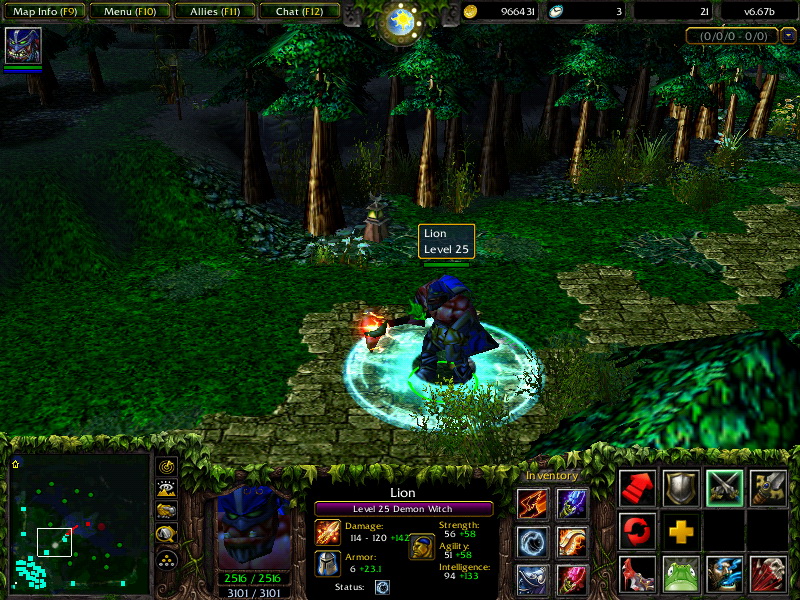}
\caption{Screenshot of the hero during a match.}
\label{fig:hero2}
\end{figure}

Each hero has distinct characteristics, behaviors and abilities. Thereby, to better focus on the strategies and the development of abilities, we designed our artificial intelligence agent to control one specific hero. The selection performed was random and the chosen character is \textit{Lion - The Demon Witch}. Given this choice, it became possible to classify which abilities and behaviors should be implemented so that the artificial intelligence agent would work with a consistent behavior during the game match. Figure \ref{fig:hero2} shows a screenshot of the character \textit{Lion - The Demon Witch} during a game match.

The DotA game also offers a variety of game modes, selected by the game host at the beginning of the match. The game modes dictate the difficulty of the scenario, as well as whether people can choose their hero or are assigned one randomly. Many game modes can be combined, allowing more flexible options. In this work we restrict the game mode to single selection, that means that a player does not receive a random hero, but is allowed to select among nine others, because Lion is automatically picked by the AI agent.

\subsection{The Challenges of Developing a MOBA agent}

Developing agents capable of defeating competitive human players in MOBA games remains an open research challenge. According to \cite{buro2003real} and \cite{weber2010applying}, improving the capabilities of computer opponents would increase the game playing experience and provide several interesting research questions for the artificial intelligence community. However, developing an AI agent to play MOBA games is not a simple task \citep{nascimento2015a}. Although there are several AI agents for all the different MOBA distributions, none of them can perform as well as expert human players. One of the reasons for this is due to the inability of AI systems to learn from experience. Human players only need a couple of matches to identify opponents' weaknesses and use them in their favor in upcoming games. Current machine learning approaches in this area are not good enough when compared to expert humans skills \citep{buro2003real, weber2011building}. Yet, according to \cite{buro2004call}, some commercial game AI systems may outperform human players and may even create challenging encounters, but they do not advance our understanding on how to create intelligent entities, since it cheats to compensate its lack of sophistication by using map revealing and faster resource gathering.

Since MOBA games are originated from Real-Time Strategy (RTS) genre, many of the challenges that surround RTS games can also be applied to MOBA. A case study for real-time AI problems in the context of RTS games can be found in \cite{buro2003real, buro2004call, buro2004rts}. 

As discussed before, MOBA provides a complex environment, populated with dynamic and static features. Moreover, MOBA characteristics tends to make the game more dynamic than its precursor, the RTS genre. Fights, duels, and actions happen in a short time, all requiring the computation of complex algorithms to analyze the scenario and to reason about it. For instance, teamfights normally last a few seconds and the agent has to perform a large amount of computation in a short time, to reason about allies, enemies, and strategies. Even for humans, it is difficult to maintain the total control of the situation during these fights.

Although not having the macromanagement that occurs in RTS games, MOBA matches require the player to reason about thousands of combinations of spells and items. The spell leveling order, item buying and building order matter, because each spell and each item has its own characteristics, making a special ability that highlights the hero early in the game. Such combinations should take into account the enemy that is being faced by the agent, the opponent team in general, the items and combinations from its own team, among many other features in the game. Even more, those features are not aways clear to be translated in a language that can be easily understood by the agent, since it requires experience and sometimes knowledge that goes beyond the game itself.

Being a commercial game genre, MOBA provides a rich hero pool, allowing the player to choose among hundreds of heroes. Performing combinations of heroes on the team can lead to success or defeat even in the hero picking phase of the game. Selecting the right hero to be played against another hero, or a set of heroes that can face the opponent's set is a difficult task. This choice requires knowledge about the teammates' heroes, the development curve, the hero classification and trying to predict the enemy's team strategy. Moreover, each hero in a MOBA game is designed with a role. That means that a hero will be better developed if it is played in the role that it was designed. Picking the right heroes for the right roles requires all the knowledge cited above, and it is a hard task for the AI agent, since it requires knowledge that goes beyond of the game scope, commonly denominated {\em metagame}. For instance, in a situation where the team composed by five weak, low-damage dealers are fighting against a team composed by five strong, high-damage dealers sounds like a bad choice, since the first team will struggle on all battles against the opponent during the match.

Lastly, MOBA games, as RTS, provide a partially observable environment. Dealing with the uncertainty of this situation is hard for most agents, because it requires sophisticated motion planning algorithms, and real-time reasoning about the environment. There are some MOBAs, like {\em Heroes of the Storm}, that even integrate a {\em bush} in the game scenario, providing spots where the player cannot be seen if his/her hero is inside a bush. This allows players to perform a wide range of tactical plays, like traps, faking and ganking. Reasoning about these fast-paced plays is not trivial, and therefore, requires predictions and especial research efforts.

\section{Methodology}

Our difficulty adjustment mechanism consists in the development of three different levels of artificial intelligence that will be chosen during the match in order to present challenges that suit the player's skills. To select the right opponent, a difficulty evaluation is performed during the game and if it indicates that the players are not evolving in the same pace, it executes the necessary adjustment. Throughout this section, we address the artificial intelligence agent developed, the game features, the difficulty evaluation process, and the mechanism to dynamically adjust the presented difficulty during a match.



\subsection{Artificial Intelligence Agent}

To be able to provide an opponent that can face different skilled players, the artificial intelligence agent must be implemented with distinct ability levels to simulate the most different behaviors played. Since the agent must simulate an opponent player, the developed algorithm implements actions and behaviors to a hero unit. During a game match, this hero should follow the player's performance, so if the player is having a good evolution, the hero controlled by artificial intelligence must be able to also do the same. However, if the player is not evolving enough or if his/her development start to decrease, the AI agent that controls the hero must lower its pace and keep up with its opponent. 

The hero behavior was divided into three categories: easy mode, regular mode and hard mode. Each one of these categories has singular aspects that aim to be suitable to players with different abilities. These are described below.

\paragraph{Easy Mode}
In the easy mode, the hero performs regular attacks every time an enemy enters in its attack range. When an allied tower is under attack, the hero detects the need for defense and moves towards the attacked ally in order to defend it. Another strategic action is how the hero chooses the enemy tower to be its main target. Every time the hero starts a moving action, it analyses which of the enemy's towers has taken more damage and is closer to be defeated. Once it finds, the hero sets that tower as the main target and goes in that direction. It is important to mention that, in the easy mode, all the attacking actions that the hero performs are basic attacks. The hero also retreats as a defense strategy. So when its health points are below 30\%, it starts to retreat towards its base, where it can recover its health when it reaches a specific recovery building. The easy mode was created for beginners or some less skilled players, where the implemented strategies are not very complex and do not use any special character skill (also known as spells). 

\paragraph{Regular Mode}
In the regular mode, besides the strategies implemented for the easy mode, the hero also starts to manipulate items. The item manipulation is very helpful to improve the hero's attributes and also to recover some attributes that have been decreased, for example, items to recover health points or mana. Likewise, there are items to increase attributes like strength, speed, intelligence, among others. As part of the defense strategy, if the hero's health points reach 30\% or less, it will first use some health potion to recover it and if these items are over, then the hero starts to retreat towards its base. The regular mode was created to match those players that have already some experience and know how to use some of the game functionalities in his/her favor but are not experts yet.

\paragraph{Hard Mode}
The hard mode has all the strategies implemented on both preceding modes, besides its own specific actions. Here, the hero goes beyond item manipulation and starts to learn, improve and cast spells. Spells are unique skills that each hero has. These spells can give a more effective damage on the enemy, can boost the recovery of its own attributes (like mana or health points), can give some kind of advantage to allied units (like freezing the enemies), among other possibilities. Every time the hero gains a new level it also gains one attribute point to distribute among its spells. So in this mode, besides the regular attack, the hero also casts spells to attack enemies or defend allies. Here we also decided to implement a new strategy for a head-to-head combat. In order to avoid losing the combat against another hero, the artificial intelligence agent algorithm keeps monitoring the area around its hero. Therefore, if an enemy hero enters the monitored area, the hero controlled by the agent will take advantage on that and will begin to attack it. The strategies to defend allied towers and to retreat are the same developed on regular mode. The hard mode was created to match those players that have more experience on the DotA game and also know how to use the game functionalities in their favor. This kind of player may be an expert on the game or a quick learner.

\begin{figure}
\centering
\includegraphics[width=12cm]{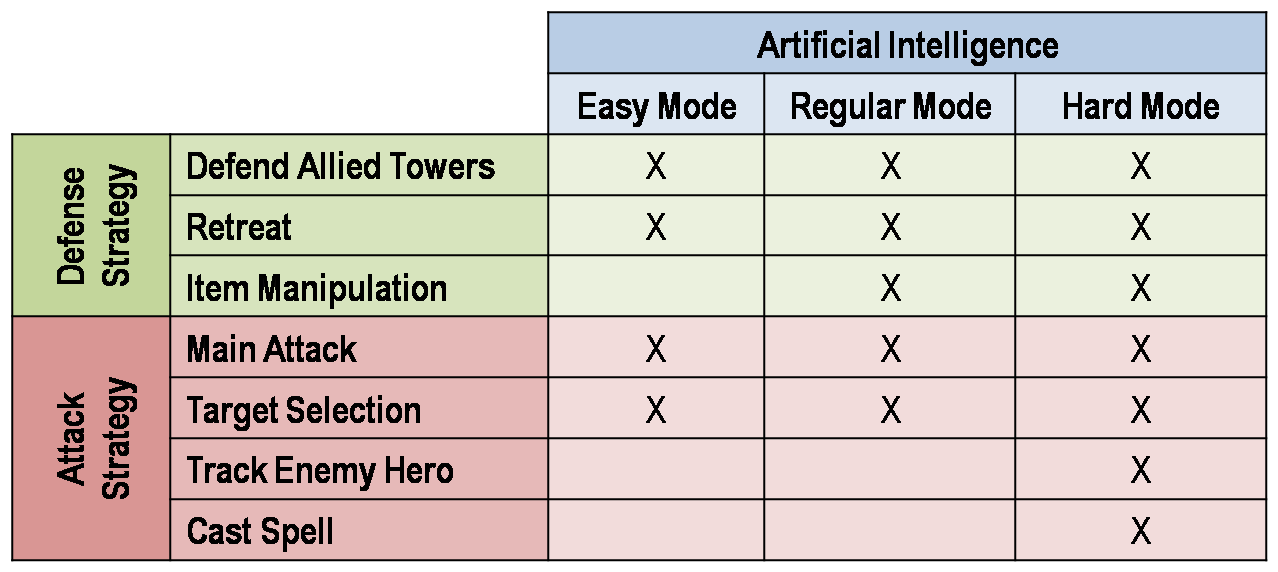}
\caption{Summary of the developed strategies for each artificial intelligence agent.}
\label{fig:tabelaIAs}
\end{figure}

The table displayed in Figure \ref{fig:tabelaIAs} summarizes all the developed difficulty modes and their strategies.

\subsection{Difficulty Evaluation Process}

A difficulty evaluation process was developed to be performed during the game and indicates when the players are not evolving at the same pace. For that, it was necessary to observe which game features should be analyzed and how to properly use the information from each one of them. The analyzed features and the evaluation process are described below.

\subsubsection{Game Features}

To evaluate a game match, it is crucial to identify which features can represent the players' performance and are relevant to the evaluation. In our testbed, we identified three important features that can illustrate the player's behavior during a DotA match. These features are: Hero's Level, Hero's Death and Towers Destroyed. Each one of these features will be described in the following paragraphs:

\paragraph{Hero's Level}
This feature represents the player's evolution during a match, where the greater is the level value, the stronger is the character. Although this feature represents the evolution, it should not be the only analyzed feature because it is possible that the player increases his/her hero's level without really increasing his/her abilities. For example, the player can keep the hero closer to battles without engaging in any fight and, by doing that, it will gain some experience points that are shared among the allies that are closer to the battle and will help the hero to evolve its level. Thereby, even if all players have heroes with equivalent levels, this feature alone does not give a real track on the game balance.

\paragraph{Hero's Death}
This feature counts how many times the hero has died during a DotA match. Differently from all other features, the hero's death may represent the player's performance and the level of difficulty that he/she is facing more accurately. For example, an inexperienced player, even having a hero with a high level, may have a high death rate, since he/she may not know how to use more properly the characteristics and peculiarities of his/her character as well as a possible lack of game strategies. Thereby, this feature seems to represent more accurately how well the player is facing the game challenges.

\paragraph{Towers Destroyed} 
This feature is the amount of enemy's towers destroyed by the allied team. It represents the team expansion and dominance over the map. Although this feature is not directly related to the player's performance, since other allies can also destroy towers, it gives us a good notion of the game's progress and team expansion over the map. Therefore, if a team is quickly progressing over the map, it may represent that the game is unbalanced.

\subsubsection{Tracking Player's Performance}

In order to perform a dynamic difficulty adjustment, it is necessary to evaluate the game from time to time and verify if the game is presenting challenges suitable to the player's performance. If the player is having a poor performance, the game should be capable to identify that and reduce its difficulty. In the same way, if the player evolves faster than the challenges presented, the game should increase its difficulty.

Once we have defined the game features that must be analyzed, this process can be summarized into the creation of an heuristic function that will keep track on the player's performance and inform when it is necessary to adjust the difficulty. This heuristic function will be our evaluation method during the game match and from now on it will be called as evaluation function. So, considering the features mentioned before and the impact that each one represents on the player's performance, we have:
\begin{equation}
P(x_t) = H_l - H_d + T_d,    
\end{equation}
where $P(x_t)$ is the performance function of player $x$ on time $t$. $H_l$ is the hero's level, $H_d$ is the hero's death count and $T_d$ is the number of towers destroyed. It is important to mention that the values of these features are related to the player and his/her hero. Computing the difference between the measurements at two consecutive times $t$ and $t-1$, it is possible to calculate the current evolution of the player, as shown in the equation below:
\begin{equation}
P'(x) = P(x_t) - P(x_{t-1}).
\end{equation}

Once the performance function was calculated for both players ($x$ and $y$) the evaluation value can be obtained by:
\begin{equation}
\alpha = P'(x) - P'(y),
\end{equation}
where $\alpha$ is the difference between performances. We should mention that player $x$ is the one that we are analyzing and player $y$ is the one controlled by the artificial intelligence system. Therefore, the player $y$ is the one that will have its difficulty adjusted during the game. It is important to measure the opponent's performance in order to evaluate if its progress is compatible or not with the players.

\subsection{Dynamic Difficulty Adjustment Mechanism}

The proposed mechanism is the key to make the adjustment work properly during the game. Until now we have only showed how to verify if the player's performance is balanced to a certain opponent or not. Thereby, the main task of the implemented mechanism is to analyze the $\alpha$ value and perform or not the difficulty adjustment at the game time $t$. 

\begin{figure}
\centering
\includegraphics[width=10cm]{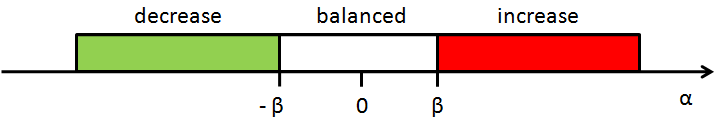}
\caption{The verification performed by the adjustment mechanism.}
\label{fig:limites}
\end{figure}

The mechanism works by evaluating the $\alpha$ variable and constantly verifying if this variable is within the $\beta$'s range, where $\beta$ represents the limit value of the evaluation function. This value means how far a player can perform better than the other player, without considering the game unbalanced. If the value of $| \beta |$ is a large number, then the adjustment will occur with less frequency, since it may take some time to $\alpha$ overcome $\beta$. Likewise, if $| \beta |$ is a small number, then the adjustment will occur more frequently, since it may overcome $\beta$ more easily. And if $\alpha$ stays inside the limits values of  $-\beta$ and $\beta$, it means that both players are having a similar performance and therefore, the match is currently balanced. The Figure \ref{fig:limites} illustrates this approach. In Section 5, several experiments were made in order to find the best limit value for $\beta$.
\section{Experiments: Agents vs Agents}

In order to verify the effectiveness of the proposed mechanism, a series of experiments was performed. The players' performance were analyzed along with the behavior of their heroes. The dynamic adjustment mechanism was also observed, as well as its variations and the impact caused on the matches.

On each experiment, we performed 20 matches of the game with the static artificial intelligence agent controlling one team against the dynamic artificial intelligence agent controlling the other one (Figure \ref{fig:tab-vit}, in the end of this section, shows a summary of the matches). The $\beta$ limits for triggering the artificial intelligence switch were set to $-1$ and $1$. Therefore, every time the difference among performances ($\alpha$) exceeds the $\beta$ limits, the difficulty of the dynamic AI should be modified accordingly. These values were determined empirically, after executing a large number of tests and observing the results contained in the gamelogs, and were not changed during these experiments.

\subsection{Baseline}

\begin{figure}[h]
\centering
\includegraphics[width=8.5cm]{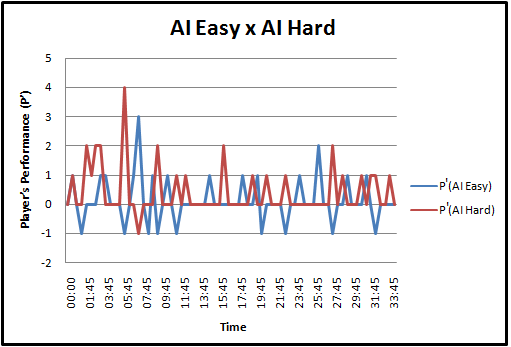}
\caption{Baseline values obtained from a match with static difficulty for both players.}
\label{fig:grafico0}
\end{figure}

First, we performed an unbalanced match in order to stipulate a baseline to compare with the obtained results from all three experiments. This baseline match is set by two different AI agent players with static behavior. One of them is on easy mode, representing a player without experience, and the second one is on hard mode, representing a very experienced player. The results of this match are shown in Figure \ref{fig:grafico0}, where the difference among both performances can be noticed. 

The player / agent performance is measured taking into account its current state during the match. The positive peaks represents moments where the agent improved its performance when compared to its last state. Likewise, negative peaks mean that the agent had its performance decreased based on its last game state. Converting these to game situations, when a hero gains a level or the team manage to destroy a tower, then this will impact positively in its development, increasing the player's current performance. Similarly, if a hero dies this will result in a negative impact in its development decreasing the player's current performance.

During this game match, the hard mode player kept increasing his performance, presenting only one time of regression in his development. Meanwhile, the easy mode player performance was very unstable, with several moments of regression in his development. Therefore, we can consider that a match will be balanced if the difference among both performances were not divergent. So, examining once again the graph of Figure \ref{fig:grafico0}, it is possible to observe that each performance peak shows itself as an appropriate moment to execute a difficulty balance in order to get the players' performance closer to each other. 

\begin{figure}[h]
\centering
\includegraphics[width=8.5cm]{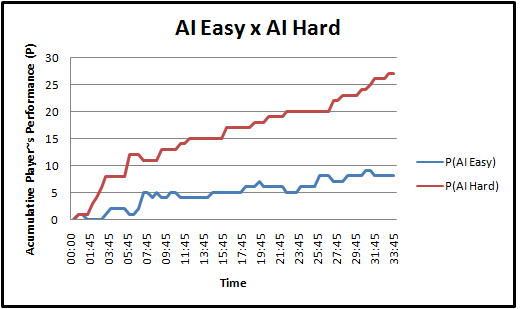}
\caption{Cumulative player's performance with static difficulty for both players.}
\label{fig:grafico0-acumulative}
\end{figure}

Figure \ref{fig:grafico0-acumulative} shows the cumulative performance value for each player during this particular match. On this graph, it becomes clear that the hard mode player evolves much faster than the easy mode player. This greater performance evolution can be related to the fact that the hero increases his level rapidly and has a low amount of deaths. On the other hand, the easy mode player had his performance lowered by a large number of deaths, in spite of having a good development.This led to a poor performance when compared to the hard mode player. 

Therefore, due to that difference between them, the adjustment appears to be necessary in order to minimize this disparity among their behaviors and present a more fair and competitive game.

After setting a baseline of an unbalanced match, we performed a set of experiments to compare the performance of a player using the adaptive AI against an opponent using a static AI, fixed at one of the predetermined modes. Specifically, for player A we used a static artificial intelligence agent in order to simulate the possible skills of a human player. For player B we applied the proposed mechanism, so that this player should keep its performance equivalent to player A and for that it should perform a dynamic difficulty adjustment. These experiments are mentioned on the following subsections.

\subsection{Easy x Adaptive}

\begin{figure}[h]
\centering
\includegraphics[width=8.5cm]{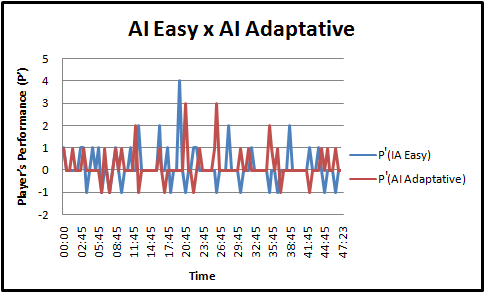}
\caption{Performance of Easy x Adaptive players during one match.}
\label{fig:grafico1a}
\end{figure}

\begin{figure}[h]
\centering
\includegraphics[width=8.5cm]{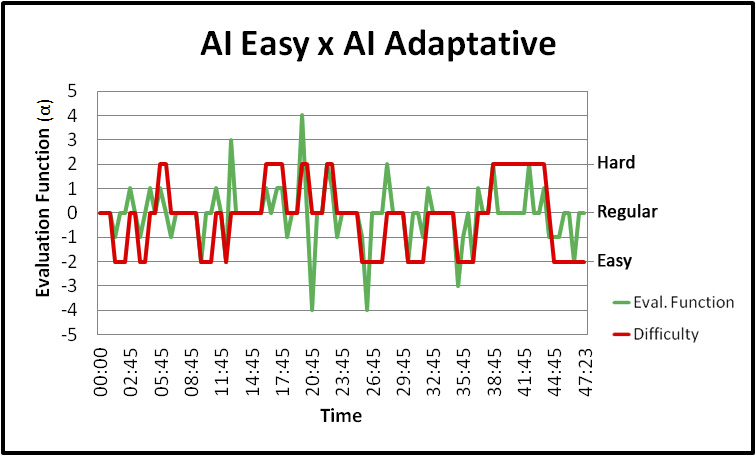}
\caption{Difficulty adjustments performed by the mechanism during one match.}
\label{fig:grafico1b}
\end{figure}

In this first set of experiments, player A was set with the easy mode, simulating a novice player, while player B was set with the adaptive AI, starting with the regular mode and switching to try to match the other player performance. Figure \ref{fig:grafico1a} shows the performance of both players during this match ($P'$), while Figure \ref{fig:grafico1b} shows the results of the evaluation function ($\alpha$) and the difficulty adjustments made during the game.

As mentioned before, the player's performance is measured by taking into account his/her current state during the match. The positive peaks represent moments where the player had improved and negative peaks mean that the player had decreased based on his/her last game state. Figure \ref{fig:grafico1a} shows that the adaptive artificial intelligence agent (player B) managed to keep its performance similar to its opponent, the easy mode player A.

On Figure \ref{fig:grafico1b}, we can track how well the adaptive player (player B) managed to be compatible with player A during the match. When the evaluation function shows negative peaks, it means that the difficulty should be adjusted and decreased by one level. Likewise, if there are positive peaks, the difficulty of the adaptive player should be increased by one level. Moments where the evaluation function remains constant (equals 0) means that the performance of both players are very similar and due to that no adjustment is necessary at this time. Therefore, the difficulty can be maintained.

It is important to mention that the difficulty adjustment is performed by increasing or decreasing one level of each time. With this approach we minimize the possibility of the opponent player noticing the behavior change. After analyzing this set of experiments and studying the gamelogs obtained from each one, we observed that in 85\% of the matches, the adaptive player B managed to keep the game balanced and as result of each match, player A won 60\% of the matches and player B won 40\%.

\subsection{Regular x adaptive}

\begin{figure}[h]
\centering
\includegraphics[width=8.5cm]{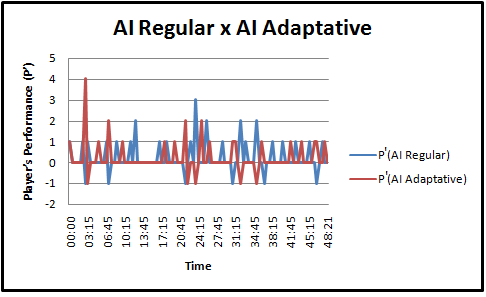}
\caption{Performance of Regular x Adaptive players during one match.}
\label{fig:grafico2a}
\end{figure}

\begin{figure}[h]
\centering
\includegraphics[width=8.5cm]{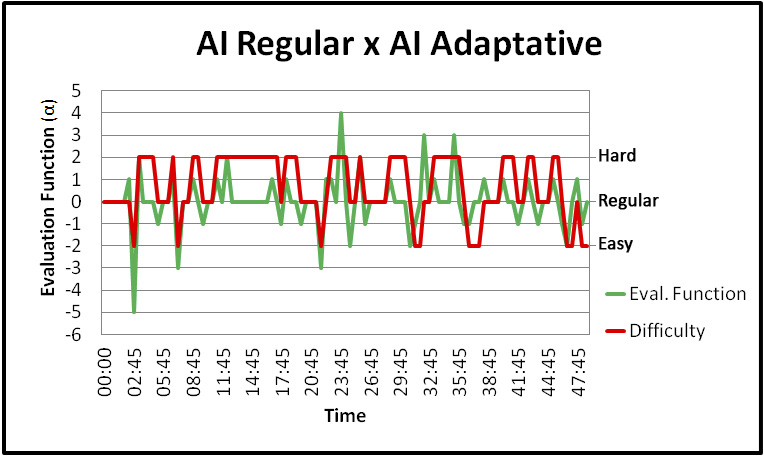}
\caption{Difficulty adjustments performed by the mechanism during one match.}
\label{fig:grafico2b}
\end{figure}

On the second set of experiments, we kept using the artificial intelligence agents  developed to control two players, one from each team. Here, we manage to simulate an intermediary player with player A using a static artificial intelligence agent on regular mode. For player B we applied the proposed mechanism, starting it on regular mode. Figure \ref{fig:grafico2a} shows the performance of both players ($P'$) during one single match. Likewise, Figure \ref{fig:grafico2b} shows the results of the evaluation function ($\alpha$) during the game and the difficulty adjustments made along the match.

The analysis performed in this set of experiments is pretty similar to the previous one. The positive peaks represent moments where the player had improved and negative peaks mean that the player had decreased its performance. In Figure \ref{fig:grafico2a}, we can observe that the adaptive artificial intelligence agent (player B) tried to follow its opponent's performance (player A) presenting similar peaks at close time periods.

On Figure \ref{fig:grafico2b}, we can follow all the adjustments made during the match. The adaptive player spent most of its time alternating between the regular mode and the hard mode. This variation can be understood as moments where player B was having a poor development when compared to player A, and the need to increase the difficulty was perceived. Similarly, when player's B behavior were standing out, the need for reducing the difficulty could also be seen. The graphic also shows that player B stayed balanced during the game. Furthermore, after analyzing this second set of experiments and studying all gamelogs collected, we observed that the players had a compatible performance in 90\% of the matches. The results of the matches were 50\% of victories for each player.

\subsection{Hard x adaptive}

\begin{figure}[h]
\centering
\includegraphics[width=8.5cm]{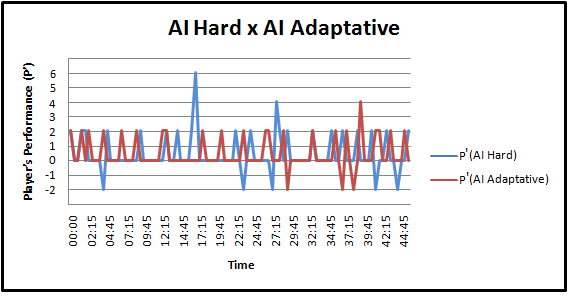}
\caption{Performance of Hard x Adaptive players during one match.}
\label{fig:grafico3a}
\end{figure}

\begin{figure}[h]
\centering
\includegraphics[width=8.5cm]{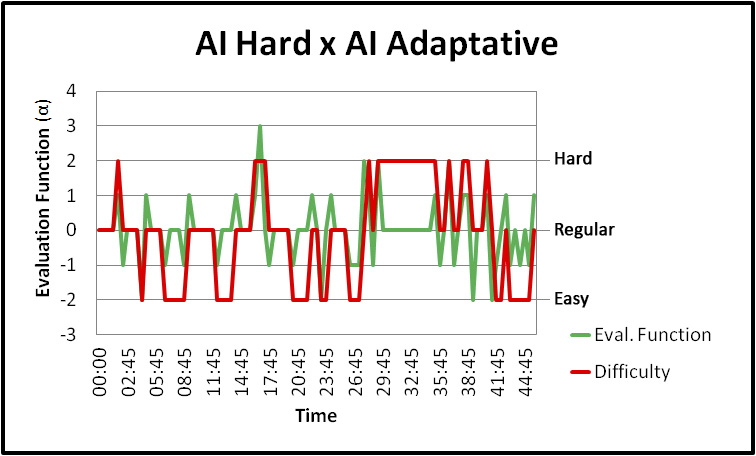}
\caption{Difficulty adjustments performed by the mechanism during one match.}
\label{fig:grafico3b}
\end{figure}

On the last set of experiments, we managed to simulate an expert player (player A) against the adaptive player (player B). As we mentioned before, the adaptive player started on regular mode and changed its behavior during the match in order to keep the game balanced. Figure \ref{fig:grafico3a} shows the performance ($P'$) of both players during one match. Likewise, Figure \ref{fig:grafico3b} shows the results of the evaluation function ($\alpha$) during the game and the difficulty adjustments made along the match.

Analyzing the results from Figure \ref{fig:grafico3a}, the adaptive player started developing a better performance than player A in the beginning of the match. Therefore, it was detected that the difficulty should be reduced in order to keep the balance (Figure \ref{fig:grafico3b}). After that, they kept their performances very close and the difficulty kept alternating between easy mode and regular mode until player A can present itself better/stronger than player B. The opposite can also be seen, when player B kept alternating between regular mode and hard mode in order to reach player's A performance.

Furthermore, after analyzing the gamelogs collected from each game, we observed that the adaptive player (player B) changed its difficulty and succeed to keep the match balanced on 80\% of the experiments. As result of the battles, player A won 45\% of the matches.

\subsection{Discussion}

Despite the good results, not all the cases presented the expected results, which has resulted in unbalanced matches. To get to this conclusion, we observed all the executed matches and studied all the collected gamelogs. These gamelogs kept track of the game on every 15 seconds, recording the current situation of both teams, the related features, their values, among other information. Once the game was finished, we started to translate those collected information, comparing the values from both heroes and making the necessary assumptions.

Considering all the performed experiments 10\% of them were unbalanced because the mechanism took too long to perform each adjustment, leading to a great difference between the players performance. So, when the players were getting closer to a balance, the match has ended. On the other hand, 5\% of the executed experiments were unbalanced due to an excess of adjustments. In these scenarios, the adjustments were being performed too quickly, causing player B to not evolve properly during the match, which resulted in an easy game for player A. 

\begin{figure}[h]
\centering
\includegraphics[width=6.5cm]{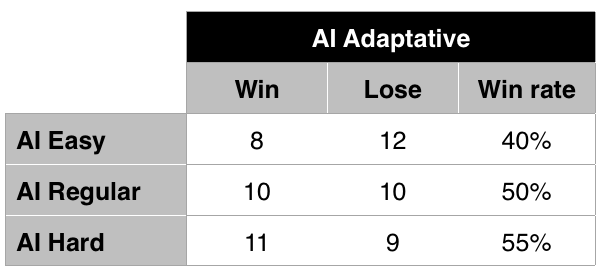}
\caption{Final results from all matches performed on the experiments.}
\label{fig:tab-vit}
\end{figure}

After performing all the experiments, it was possible to summarize the obtained final results from the game matches. Figure \ref{fig:tab-vit} shows the amount of victories and losses of the adaptive AI against the easy, regular and hard modes. These results show that both players had a similar amount of victories, demonstrating that the adaptive player offers a more balanced game experience.

\section{Experiments: Agents vs Users}

We also performed tests with users to assess qualitatively the efficacy of the implemented mechanism. The objective of these tests consists of verifying, according to the perception of the player, if the mechanism can keep the difficulty of the game balanced facing their skills and if this approach stimulates his/her entertainment. To make the tests more objective, we only enrolled users who have played the game \textsc{Defense of the Ancients} (DotA) at least once, in order to avoid problems in understanding interface elements and/or the game's mechanics.

The tests involved inviting the users to play two matches of the game \textsc{Defense of the Ancients} (DotA), where their main goal was to defeat the opposing team. However, it was explained to them that the outcome of the match was not crucial for the experiment, since we were more interested in behavioral issues generated by the game. We provided two types of maps. On map A, users faced the dynamic agent as opponent and on map B they faced a static agent. With these two types of maps, we evaluated if the difficulty adjustment can be perceived during the match and whether or not this dynamic adjustment can really impact on the player entertainment. Note that each volunteer played two times on the same map, not changing maps. We performed the tests on this way because we were trying to avoid the placebo effect on the users. Once the user knows that he was playing in both game versions he could assume that the adjustment was happening even though it may not be true. Therefore, we managed to ask only the expert users to play against both opponents (dynamic agent and static agent), in order to really assess if the adjustment can be detected or not. We choose to ask only the expert players to play in both maps because they have more knowledge and expertise in the game mechanics, this makes them able to identify quickly any abnormal behavior on the opponent.

A total of eleven users participated on the tests. All tests were guided by the following script:

\begin{enumerate}
\item The participant was instructed about how the test was going to occur and signed an agreement to participate on the test;
\item A quick presentation of the game mechanics was carried out and its rules in case the user did not remember it;
\item A preliminary interview was made, whose main objective was to let us know the user's profile and background regarding game playing;
\item The participant played two game matches of DotA on the same map;
\item A post-test interview was made, whose objective was to gather the user's opinions and perceptions regarding the presented system.
\end{enumerate}

The results from these tests are discussed in the following sections.

\subsection{Pre-test evaluation}

During the tests, all users were instructed to answer a series of questions to make it possible to establish profiles according to their habits, preferences and experiences. These questions were placed on the pre-test questionnaire and its main goal is to try to identify similarities among the volunteers' behavior during the experiments. 

A total of eleven male users participated on the experiments. Among them one is less than 18 years old, two are from 18 to 21 years old, four are from 22 to 25 years old and the last four are between 26 and 29 years old. Regarding their education level, one was still in high school, seven were studying or already concluded a higher education and one had master's degree. 


\begin{figure}[h]
\centering
\includegraphics[width=12cm]{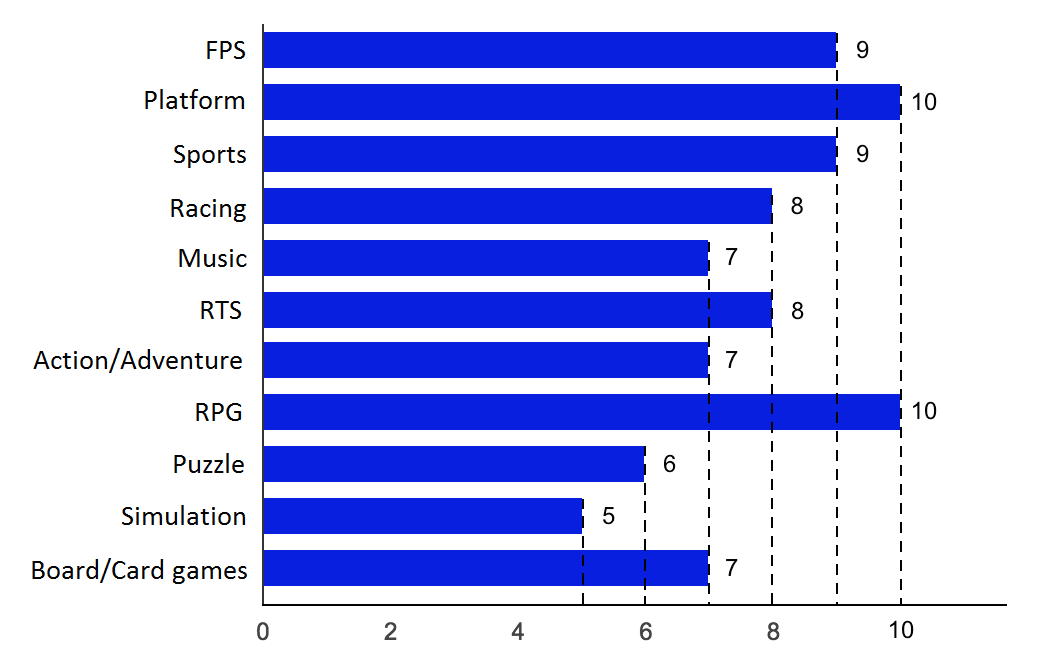}
\caption{Most popular game genres among the volunteers.}
\label{fig:genre}
\end{figure}

To verify their experience with digital games, we asked which are the game genres that the users have recently played the most. The most popular genres were platform games and RPG selected by 90.9\% of the volunteers, followed by First Person Shooter (FPS) and sports with 81.8\% of the users and on third place there is Real-Time Strategy (RTS) and racing with 72.2\%. Observe that this percentages do not sum up, because users could select more than one option. Figure \ref{fig:genre} shows all the genres selected by the volunteers.

\begin{figure}[h]
\centering
\includegraphics[width=12cm]{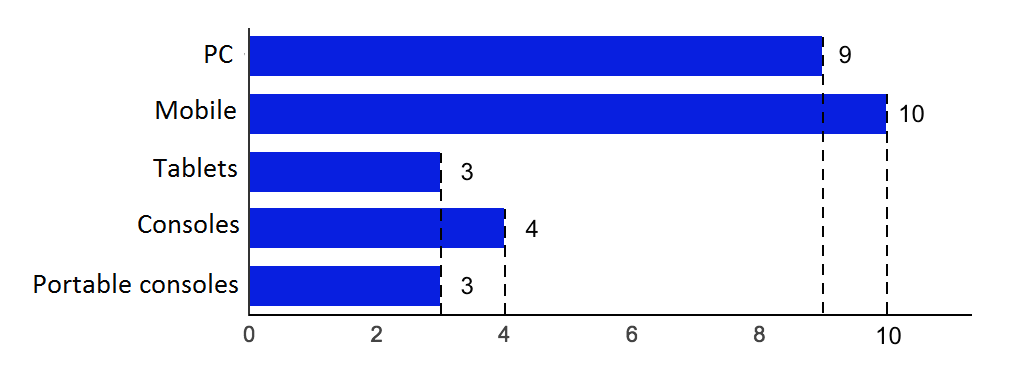}
\caption{Most popular gaming platforms among the volunteers.}
\label{fig:device}
\end{figure}

Regarding the frequency on which they play, 63.6\% of the users play every day and 36.4\% from 1 to 3 times per week. The players were also asked on which devices they are currently playing more and 90.9\% of the users selected mobile devices, followed by PC with 81.8\% and consoles with 36.4\%. Figure \ref{fig:device} presents the chart with the most popular devices.

Now, regarding MOBA games knowledge, all users said that they had played DotA at least once, on which five rated themselves as beginners, four as intermediary players and two as experts. Among the eleven volunteers, only eight said they had played another MOBA game: 100\% of them played {\em League of Legends} and 37.5\% played {\em Heroes of Newerth}. About their expertise in MOBA games in general, 25\% self-proclaimed as beginners, 37.5\% as intermediary players and 37.5\% as expert players.

\subsection{Post-test evaluation}

After answering all pre-test questions, the participants were asked to play two matches against our agents. Therefore, we created two different maps (A and B) and asked the volunteer to play both matches in only one map. Both maps contain the same game structure and what differ them is that map A hosts the dynamic artificial intelligence agent while map B hosts the static artificial intelligence agent. These agents are the same used on the previous experiment. Among the participants, six of them played on map A and five of them played on map B.

Once both matches were concluded, the user was instructed to answer the post-test questionnaire, which tackles various aspects related to how he/she perceived the game experience. Questions related to experience and immersion of the user during the match came from a selection of questionnaires about user experience in games \citep{takatalo2015understanding, fox2013development, jennett2008measuring, ijsselsteijn2007characterising}. They were presented as affirmatives so that the user could choose how much he/she agrees to it following the 5 points of {\em Likert's} classification \citep{norman2010likert}. The scale goes from 1 - ``Strongly disagree''; 2 - ``Partially disagree''; 3 - ``Indifferent''; 4 - ``Partially agree''; and 5 - ``Strongly agree''.

\begin{figure}[h]
\centering
\includegraphics[width=12cm]{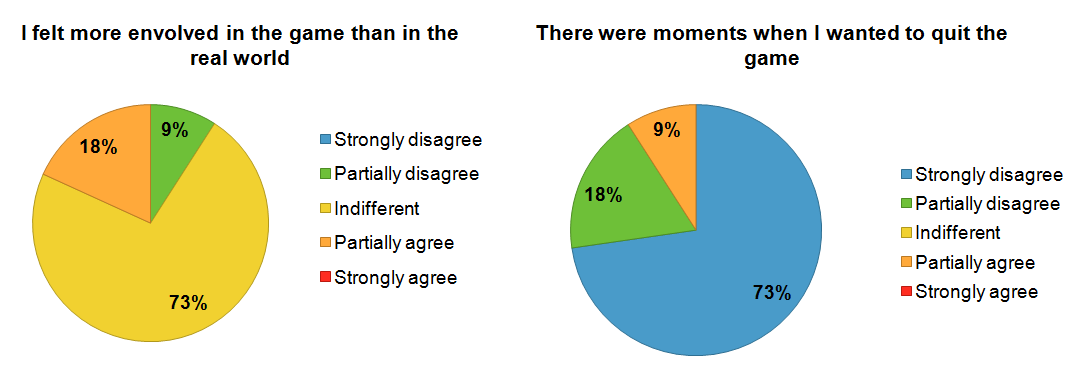}
\caption{Affirmatives that address aspects of the player immersion during the match.}
\label{fig:pizza1}
\end{figure}

The first set of affirmatives addresses aspects of the player immersion during the match: 55\% described as indifferent when asked if they did not realize the time running while playing and 73\% of them strongly agree that they worked hard to get good results in the game. When asked if there were moments where they wanted to quit the game, 73\% strongly disagree with this affirmative . This may suggest that although half of the volunteers said they were indifferent regarding the game time lapse, the majority stated that they put some effort to accomplish the main goal and they did not want to quit (Figure \ref{fig:pizza1}). Thereby, we can assume that the game was able to offer a considerable level of immersion to the player.

\begin{figure}[h]
\centering
\includegraphics[width=14cm]{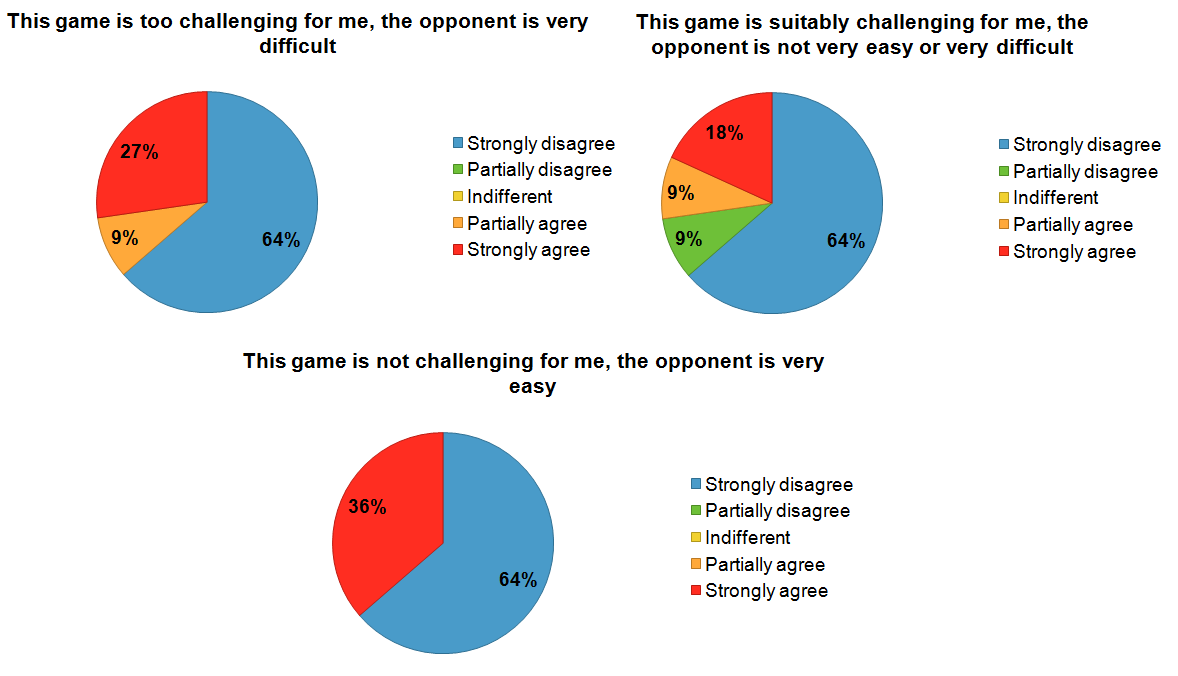}
\caption{Affirmatives that address the game challenge provided by the agent during the match.}
\label{fig:pizza2}
\end{figure}

The next set of affirmatives addresses the game challenge provided by the agent during the match. When asked if the game kept them motivated to keep playing, 18\% strongly agree and 46\% partially agree. Now, regarding the difficulty, 27\% strongly agree that the game is too challenging for them, while 18\% strongly agree that the game is suitably challenging for them. Finally, 36\% said that the game is not challenging at all (Figure \ref{fig:pizza2}). These very distinct opinions can be observed due to the different level of expertise of each player. Those that consider themselves as beginners in DotA or general MOBA, found that the game was too challenging for them. Using the same analysis, those that consider themselves as experts found the game too easy because they know more advanced strategies that goes beyond the agent's algorithm. 

\begin{figure}[h]
\centering
\includegraphics[width=14cm]{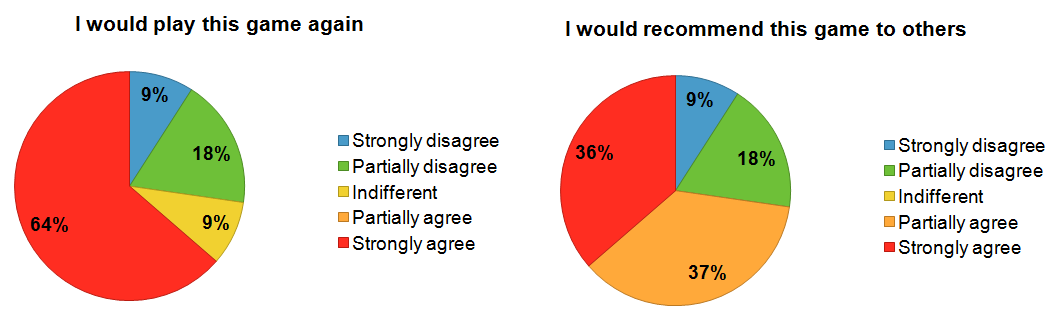}
\caption{Affirmatives that address the enjoyment during the match.}
\label{fig:pizza3}
\end{figure}

The following set of affirmatives addressed the player ability/competence during the game. 64\% of the players felt successful at the end of the game and won the match. Also, 82\% agree that they were making progress during the course of the game. About the player enjoyment during the game, 73\% liked to play against our agent and would recommend this game to others. 64\% said that he would play this game again (Figure \ref{fig:pizza3}). Therefore, although the agents were not always very challenging to all players, we can assume that most of them enjoyed the match against it.

The last set of affirmatives addresses the agent opponent level. Here, all the opinions got divided where we have that 27.3\% strongly agree and 9.1\% partially agree that the opponent plays a lot better than they. This opinion were given probably by beginners, since they may had a more difficult time trying to win the match. Regarding that the opponent plays in the same level as them, 27.3\% strongly agree with this affirmation, which may indicate that these players are intermediaries or maybe beginners that managed to win the match. Finally, we have that 36.4\% of the players strongly agree that the opponent plays much worse than them. This answer was given by experts and intermediaries players that managed to win the match without putting too much effort on this accomplishment. Thereby, we can observe that the player opinion about the opponent reflects directly on his/her expertise, where the agent with the same behavior can be too challenging for some and not very challenging for another.

To analyse the player's perception during the game, we asked them some questions about whether or not they have noticed any changes on the agent's behavior during the match. We made these questions to all the players regardless the map where they played. All of them said they did not notice the opponent getting harder or easier. They also mentioned not noticing the agent adapting its behavior in order to be more suitable to them. Therefore, based on these answers we can assume that the mechanism were successful in at least one of its goals by changing its behavior without making it noticeable to the player. This is the first premise of the dynamic difficulty adjustment which tries to guarantee that the player will not feel cheated or disappointed during the game.

\subsection{Discussion}

The main goal of these experiments were to evaluate qualitatively the efficacy of the implemented mechanism, verifying if it can keep the game difficulty balanced to each player, and whether or not this can really impact on the player entertainment. Therefore, we provided two types of maps, one with the dynamic agent and another with the static agent.

After analysing the results from all players, we asked informally for both expert volunteers to also play on the other map, so that we could assess their perceptions regarding both agents. After comparing the experience they had on each map, they said that the opponent from map A (dynamic agent) presented a more fluid behavior and they preferred to play against it. They also stated that both maps were not very challenging to them and maybe it could be more suitable for players with little experience. With that feedback we believe that, in the current state, our agent has enough expertise to play against novice players. Thereby, the novice players can be entertained while learning the game style.

By observing the data regarding the player immersion during the match, we can consider that the game provided a satisfactory level of immersion, since none of the players said they wanted to quit the game or found it too long. As well as the questions related to the player enjoyment during the match, although the agents were not always very challenging against all users, we can assume that most of them enjoyed playing against it.

However, after considering the responses related to the game challenge provided, it was observed that, for experts, the developed artificial intelligence agent turned out to be weak and not very challenging. We believed that this occurred due to the absence of more complex strategies during its development, once it is not a simple task to develop such agents. As mentioned on Section 3.6, developing agents capable of defeating competitive human players in MOBA remains an open research challenge and we can attribute our agents flaws to this. Some of the intermediary players also found both agents not very difficult to defeat. As for the entry-level players, based on their responses we could state that they found both opponents too challenging, because probably they had a more difficult time trying to win the match although the agent had kept the same pace as his.

\section{Conclusion}

The dynamic difficulty adjustment consists in an alternative towards the definition of the game challenge levels. This adjustment is dynamically performed, making it possible to track the player's skills and adjust itself during game runtime.

The presented work aimed to increase the player's entertainment by providing a mechanism that adjusts the game AI agent according to the player's skills. This mechanism was implemented on a MOBA game, called \textsc{Defense of the Ancient} (DotA). After performing experiments that simulate the three main player's behaviors (beginner, regular and experienced), it was possible to verify that the dynamic difficulty adjustment mechanism was able to keep up with the player's abilities on 85\% of all experiments. On the remaining experiments that failed to suit the player's skill, 10\% of it occurred because the adjustment mechanism spent too much time to perform each needed adjustment which resulted in a great difference between the players performance. And the last 5\% of it occurred due to an excess of adjustments that were performed too quickly, without giving enough time to the game to evolve properly.

Given the presented results, we can conclude that the proposed mechanism behaved as expected and is capable to offer a game match compatible with the simulated player's performance. Also, after observing all obtained results, we can state that the key to a balanced game is to keep changing the difficulty of the adaptive player in order to follow the performance of the human player and avoid boredom and frustration.

As future work, the dynamic difficulty adjustment mechanism will be improved in order to decrease the amount of cases where the balance did not work properly. We believe that we must implement a general AI, capable of handling almost all heroes available in DotA, in order to offer diversity of strategies and helping the player to learn how to play against different heroes. Another improvement that could be done is testing the activation or deactivation of features, like item buy, spell casting and combos, tracking the player behavior and trying to imitate his/her knowledge.

In machine learning field, it is possible to try to learn the players preference, in order to improve the knowledge of the intelligent agent, making it to follow the player's behavior. Moreover, machine learning could track which skills the player has developed and push new knowledge into the agent in order to try to stimulate the player to explore the game and learn what the agent is doing. We could also track the player preferences in order to try to classify his/her play style and select heroes that match his/her preferences, making the learning curve smoother. Finally we want to performe qualitative tests with a larger number of players in order to get more insights about the proposed mechanism.

\section{Acknowledgements}

This work was partially supported by CAPES, CNPq and Fapemig. We would like to thank all the volunteers that participated in our qualitative tests. 

\section*{References}
\bibliographystyle{elsarticle-harv}
\bibliography{bibliography}

\begin{thebibliography}{49}
\expandafter\ifx\csname natexlab\endcsname\relax\def\natexlab#1{#1}\fi
\expandafter\ifx\csname url\endcsname\relax
  \def\url#1{\texttt{#1}}\fi
\expandafter\ifx\csname urlprefix\endcsname\relax\def\urlprefix{URL }\fi

\bibitem[{Andrade et~al.(2005)Andrade, Ramalho, Santana, and
  Corruble}]{andrade2005extending}
Andrade, G., Ramalho, G., Santana, H., Corruble, V., 2005. Extending
  reinforcement learning to provide dynamic game balancing. In: Proceedings of
  the Workshop on Reasoning, Representation, and Learning in Computer Games,
  19th International Joint Conference on Artificial Intelligence (IJCAI). pp.
  7--12.

\bibitem[{Bakkes et~al.(2009)Bakkes, Spronck, and van~den
  Herik}]{bakkes2009rapid}
Bakkes, S., Spronck, P., van~den Herik, J., 2009. Rapid and reliable adaptation
  of video game {AI}. Computational Intelligence and AI in Games, IEEE
  Transactions on 1~(2), 93--104.

\bibitem[{Bartle(2004)}]{bartle2004designing}
Bartle, R.~A., 2004. Designing virtual worlds. New Riders.

\bibitem[{Bowman and McMahan(2007)}]{bowman2007virtual}
Bowman, D.~A., McMahan, R.~P., 2007. Virtual reality: how much immersion is
  enough? Computer 40~(7), 36--43.

\bibitem[{Buro(2003)}]{buro2003real}
Buro, M., 2003. Real-time strategy games: A new {AI} research challenge. In:
  IJCAI. pp. 1534--1535.

\bibitem[{Buro(2004)}]{buro2004call}
Buro, M., 2004. Call for ai research in rts games. In: Proceedings of the
  AAAI-04 Workshop on Challenges in Game AI. pp. 139--142.

\bibitem[{Buro and Furtak(2004)}]{buro2004rts}
Buro, M., Furtak, T., 2004. {RTS} games and real-time {AI} research. In:
  Proceedings of the Behavior Representation in Modeling and Simulation
  Conference (BRIMS). Vol. 6370.

\bibitem[{Csikszentmihalyi(1991)}]{csikszent1991flow}
Csikszentmihalyi, M., 1991. Flow. HarperCollins.

\bibitem[{Csikszentmihalyi(2000)}]{csikszentmihalyi2000beyond}
Csikszentmihalyi, M., 2000. Beyond boredom and anxiety. Jossey-Bass.

\bibitem[{Csikszentmihalyi and Nakamura(2010)}]{csikszentmihalyi2010effortless}
Csikszentmihalyi, M., Nakamura, J., 2010. Effortless attention in everyday
  life: A systematic phenomenology. In: Bruya, B. (Ed.), Effortless Attention:
  A New Perspective in the Cognitive Science of Attention and Action. MIT
  Press, pp. 179--189.

\bibitem[{de~Araujo and Feij\'{o}(2013)}]{feijo2013evaluating}
de~Araujo, B. B. P.~L., Feij\'{o}, B., Oct. 2013. Evaluating dynamic difficulty
  adaptivity in shoot'em up games. In: Proceedings of the {XII} Brazilian
  Symposium on Games and Digital Entertainment - {SBGames} 2013. S\~{a}o Paulo,
  Brazil, pp. 229 -- 238.

\bibitem[{Demasi and Adriano(2003)}]{demasi2003line}
Demasi, P., Adriano, J. d.~O., 2003. On-line coevolution for action games.
  International Journal of Intelligent Games \& Simulation 2~(2).

\bibitem[{Egenfeldt-Nielsen et~al.(2013)Egenfeldt-Nielsen, Smith, and
  Tosca}]{egenfeldt2013understanding}
Egenfeldt-Nielsen, S., Smith, J.~H., Tosca, S.~P., 2013. Understanding video
  games: The essential introduction. Routledge.

\bibitem[{Fox and Brockmyer(2013)}]{fox2013development}
Fox, C.~M., Brockmyer, J.~H., 2013. The development of the game engagement
  questionnaire: A measure of engagement in video game playing: Response to
  reviews. Interacting with Computers, iwt003.

\bibitem[{Games and Games(2009)}]{riot2009lol}
Games, R., Games, W.~R., 2009. League of legends. Computer Game.

\bibitem[{Gaudiosi(2012)}]{gaudiosi2012riot}
Gaudiosi, J., 2012. Riot games' league of legends officially becomes most
  played pc game in the world. Forbes. Jul 11, 2011.

\bibitem[{Hunicke(2005)}]{hunicke2005case}
Hunicke, R., 2005. The case for dynamic difficulty adjustment in games. In:
  Proceedings of the 2005 ACM SIGCHI International Conference on Advances in
  computer entertainment technology. ACM, pp. 429--433.

\bibitem[{IJsselsteijn et~al.(2007)IJsselsteijn, De~Kort, Poels, Jurgelionis,
  and Bellotti}]{ijsselsteijn2007characterising}
IJsselsteijn, W., De~Kort, Y., Poels, K., Jurgelionis, A., Bellotti, F., 2007.
  Characterising and measuring user experiences in digital games. In:
  International conference on advances in computer entertainment technology.
  Vol.~2. p.~27.

\bibitem[{Jennett et~al.(2008)Jennett, Cox, Cairns, Dhoparee, Epps, Tijs, and
  Walton}]{jennett2008measuring}
Jennett, C., Cox, A.~L., Cairns, P., Dhoparee, S., Epps, A., Tijs, T., Walton,
  A., 2008. Measuring and defining the experience of immersion in games.
  International journal of human-computer studies 66~(9), 641--661.

\bibitem[{Johnson et~al.(2015)Johnson, Nacke, and Wyeth}]{johnson2015all}
Johnson, D., Nacke, L.~E., Wyeth, P., 2015. All about that base: differing
  player experiences in video game genres and the unique case of moba games.
  In: Proceedings of the 33rd Annual ACM Conference on Human Factors in
  Computing Systems. ACM, pp. 2265--2274.

\bibitem[{Klimmt et~al.(2009)Klimmt, Blake, Hefner, Vorderer, and
  Roth}]{klimmt2009player}
Klimmt, C., Blake, C., Hefner, D., Vorderer, P., Roth, C., 2009. Player
  performance, satisfaction, and video game enjoyment. In: Entertainment
  Computing--ICEC 2009. Springer, pp. 1--12.

\bibitem[{Koster(2010)}]{koster2010theory}
Koster, R., 2010. Theory of fun for game design. O'Reilly Media, Inc.

\bibitem[{Kwak et~al.(2015)Kwak, Blackburn, and Han}]{kwak2015}
Kwak, H., Blackburn, J., Han, S., 2015. Exploring cyberbullying and other toxic
  behavior in team competition online games. Social Dynamics 22~(28), 47.

\bibitem[{L{\o}voll and Vitters{\o}(2014)}]{lovoll2014can}
L{\o}voll, H.~S., Vitters{\o}, J., 2014. Can balance be boring? a critique of
  the “challenges should match skills” hypotheses in flow theory. Social
  indicators research 115~(1), 117--136.

\bibitem[{Machado et~al.(2011)Machado, Fantini, and
  Chaimowicz}]{machadoplayerA}
Machado, M.~C., Fantini, E.~P., Chaimowicz, L., 2011. Player modeling: Towards
  a common taxonomy. In: Computer Games (CGAMES), 2011 16th International
  Conference on. IEEE, pp. 50--57.

\bibitem[{Malone(1981)}]{malone1981toward}
Malone, T.~W., 1981. Toward a theory of intrinsically motivating instruction*.
  Cognitive science 5~(4), 333--369.

\bibitem[{Mateas(2002)}]{mateas2002interactive}
Mateas, M., 2002. Interactive drama, art and artificial intelligence. Ph.D.
  thesis, Carnegie Mellon University, Pittsburgh, PA, USA, aAI3121279.

\bibitem[{Missura and G{\"a}rtner(2009)}]{missura2009player}
Missura, O., G{\"a}rtner, T., 2009. Player modeling for intelligent difficulty
  adjustment. In: Discovery Science. Springer, pp. 197--211.

\bibitem[{Nacke and Lindley(2008)}]{nacke2008flow}
Nacke, L., Lindley, C.~A., 2008. Flow and immersion in first-person shooters:
  measuring the player's gameplay experience. In: Proceedings of the 2008
  Conference on Future Play: Research, Play, Share. ACM, pp. 81--88.

\bibitem[{Nareyek(2004)}]{nareyeka}
Nareyek, A., 2004. {AI} in computer games. Queue 1~(10), 58.

\bibitem[{Norman(2010)}]{norman2010likert}
Norman, G., 2010. Likert scales, levels of measurement and the “laws” of
  statistics. Advances in health sciences education 15~(5), 625--632.

\bibitem[{Poole(2004)}]{poole2004trigger}
Poole, S., 2004. Trigger happy: Videogames and the entertainment revolution.
  Arcade Publishing.

\bibitem[{Silva(2015)}]{silva2015a}
Silva, M.~P., 2015. Intelig{\^e}ncia artificial adaptativa para ajuste
  din{\^a}mico de dificuldade em jogos digitais. Master's thesis, Universidade
  Federal de Minas Gerais (UFMG), Belo Horizonte/MG, Brazil.

\bibitem[{Silva et~al.(2015)Silva, Silva, and Chaimowicz}]{silva2015b}
Silva, M.~P., Silva, V. d.~N., Chaimowicz, L., 2015. Dynamic difficulty
  adjustment through an adaptive {AI}. In: Brazilian Symposium on Games and
  Entertainment (SBGames), 2015 SBC 14th. pp. 52--59.

\bibitem[{Silva and Chaimowicz(2015{\natexlab{a}})}]{nascimento2015a}
Silva, V. d.~N., Chaimowicz, L., 2015{\natexlab{a}}. On the development of
  intelligent agents for moba games. In: Brazilian Symposium on Games and
  Entertainment (SBGames), 2015 SBC 14th.

\bibitem[{Silva and Chaimowicz(2015{\natexlab{b}})}]{nascimento2015b}
Silva, V. d.~N., Chaimowicz, L., 2015{\natexlab{b}}. A tutor agent for moba
  games. In: Brazilian Symposium on Games and Entertainment (SBGames), 2015 SBC
  14th.

\bibitem[{Smith et~al.(2011)Smith, Lewis, Hullett, Smith, and
  Sullivan}]{smithtaxonomy}
Smith, A.~M., Lewis, C., Hullett, K., Smith, G., Sullivan, A., 2011. An
  inclusive taxonomy of player modeling. University of California, Santa Cruz,
  Tech. Rep. UCSC-SOE-11-13.

\bibitem[{Spronck et~al.(2006)Spronck, Ponsen, Sprinkhuizen-Kuyper, and
  Postma}]{spronck2006adaptive}
Spronck, P., Ponsen, M., Sprinkhuizen-Kuyper, I., Postma, E., 2006. Adaptive
  game ai with dynamic scripting. Machine Learning 63~(3), 217--248.

\bibitem[{Stanley et~al.(2005)Stanley, Bryant, and
  Miikkulainen}]{stanley2005evolving}
Stanley, K.~O., Bryant, B.~D., Miikkulainen, R., 2005. Evolving neural network
  agents in the nero video game. Proceedings of the IEEE, 182--189.

\bibitem[{Takatalo et~al.(2015)Takatalo, H{\"a}kkinen, and
  Nyman}]{takatalo2015understanding}
Takatalo, J., H{\"a}kkinen, J., Nyman, G., 2015. Understanding presence,
  involvement, and flow in digital games. In: Game User Experience Evaluation.
  Springer, pp. 87--111.

\bibitem[{Thompson et~al.(2015)Thompson, Parker, and
  Cox}]{thompson2015interrogating}
Thompson, P., Parker, R., Cox, S., 2015. Interrogating creative theory and
  creative work: Inside the games studio. Sociology, 0038038514565836.

\bibitem[{Togelius et~al.(2007)Togelius, De~Nardi, and
  Lucas}]{togelius2007towards}
Togelius, J., De~Nardi, R., Lucas, S.~M., 2007. Towards automatic personalised
  content creation for racing games. In: Computational Intelligence and Games,
  2007. CIG 2007. IEEE Symposium on. IEEE, pp. 252--259.

\bibitem[{Van Den~Hoogen et~al.(2008)Van Den~Hoogen, Ijsselsteijn, and
  de~Kort}]{van2008exploring}
Van Den~Hoogen, W., Ijsselsteijn, W., de~Kort, Y., 2008. Exploring behavioral
  expressions of player experience in digital games. In: Proceedings of the
  workshop on Facial and Bodily Expression for Control and Adaptation of Games
  ECAG 2008. pp. 11--19.

\bibitem[{Weber et~al.(2010)Weber, Mateas, and Jhala}]{weber2010applying}
Weber, B.~G., Mateas, M., Jhala, A., 2010. Applying goal-driven autonomy to
  starcraft. In: AIIDE.

\bibitem[{Weber et~al.(2011)Weber, Mateas, and Jhala}]{weber2011building}
Weber, B.~G., Mateas, M., Jhala, A., 2011. Building human-level {AI} for
  real-time strategy games. In: AAAI Fall Symposium: Advances in Cognitive
  Systems. Vol.~11. p.~01.

\bibitem[{Wheat et~al.(2015)Wheat, Masek, Lam, and Hingston}]{wheat2015dynamic}
Wheat, D., Masek, M., Lam, C.~P., Hingston, P., 2015. Dynamic difficulty
  adjustment in 2d platformers through agent-based procedural level generation.
  In: Systems, Man, and Cybernetics (SMC), 2015 IEEE International Conference
  on. IEEE, pp. 2778--2785.

\bibitem[{Xavier(2010)}]{xavier2010condiccao}
Xavier, G., 2010. A condi{\c{c}}{\~a}o eletrol{\'u}dica: Cultura visual nos
  jogos eletr{\^o}nicos. Teres{\'o}polis: Novas ideias.

\bibitem[{Yannakakis(2008)}]{yannakakis2008model}
Yannakakis, G.~N., 2008. How to model and augment player satisfaction: a
  review. In: Proceedings of the 1st Workshop on Child, Computer and
  Interaction (ICMI'08), ACM Press, Montreal, Canada.

\bibitem[{Yannakakis and Hallam(2007)}]{Yannakakis07towardsoptimizing}
Yannakakis, G.~N., Hallam, J., 2007. Towards optimizing entertainment in
  computer games. Applied Artificial Intelligence.

\end{thebibliography}

\newpage
\appendix
\section{User Experiments Questionnaire}

\begin{figure}[h]
\centering
\fbox{\includegraphics[width=13cm]{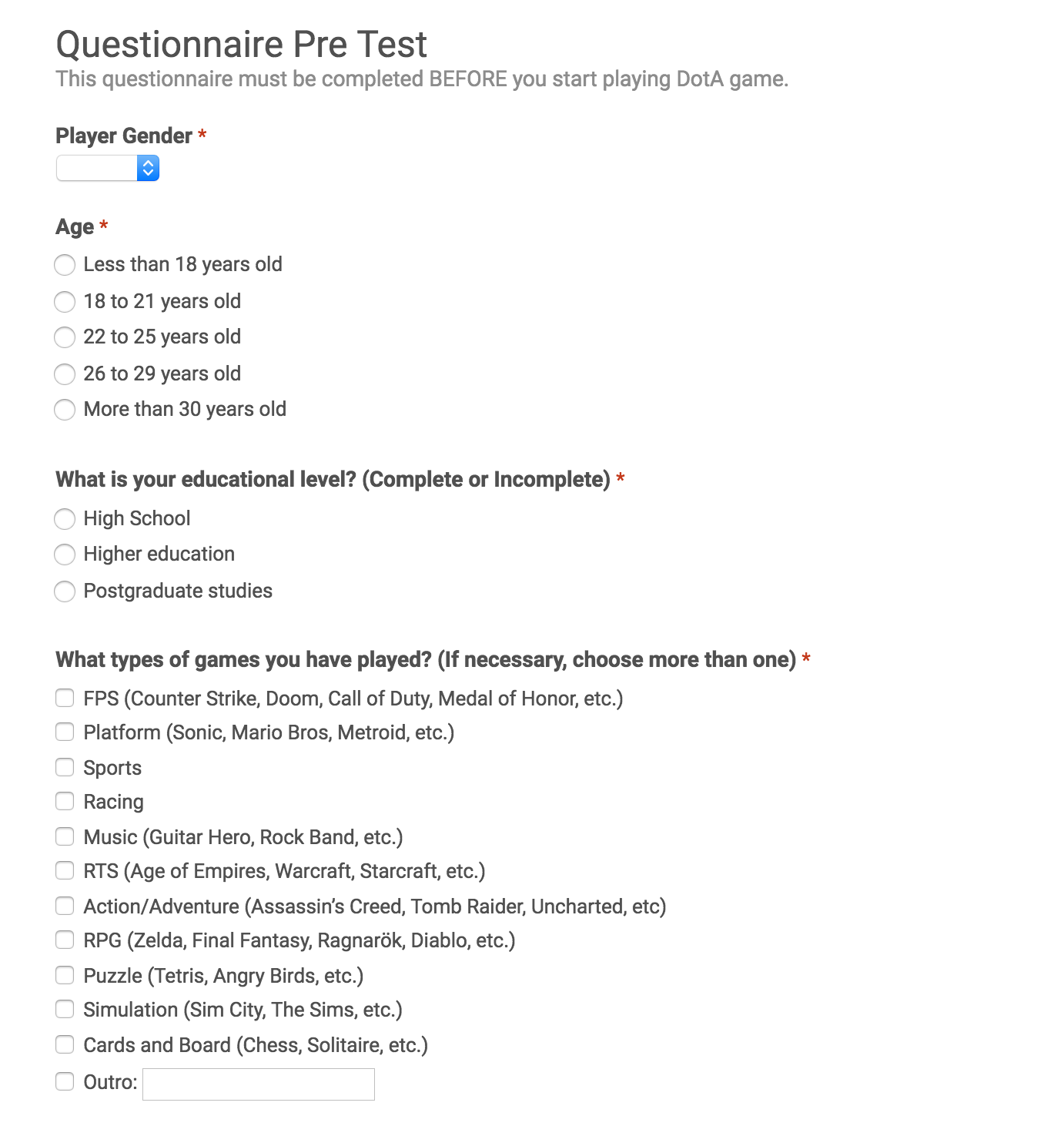}}
\caption{Pre-test questionnaire.}
\label{fig:form_cons}
\end{figure}

\begin{figure}[h]
\centering
\fbox{\includegraphics[width=13cm]{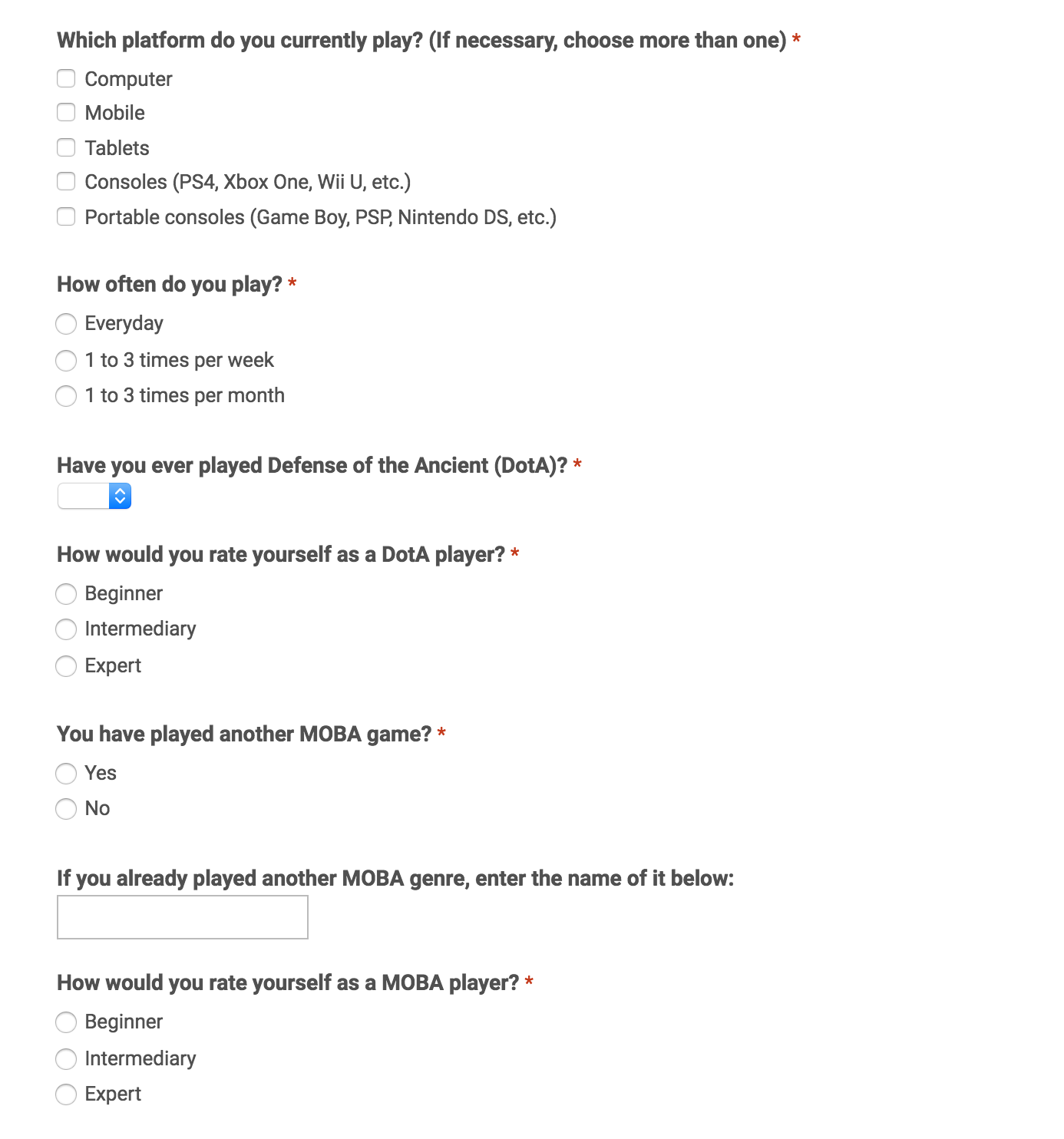}}
\caption{Pre-test questionnaire.}
\label{fig:form_pre1}
\end{figure}

\begin{figure}[h]
\centering
\fbox{\includegraphics[width=13cm]{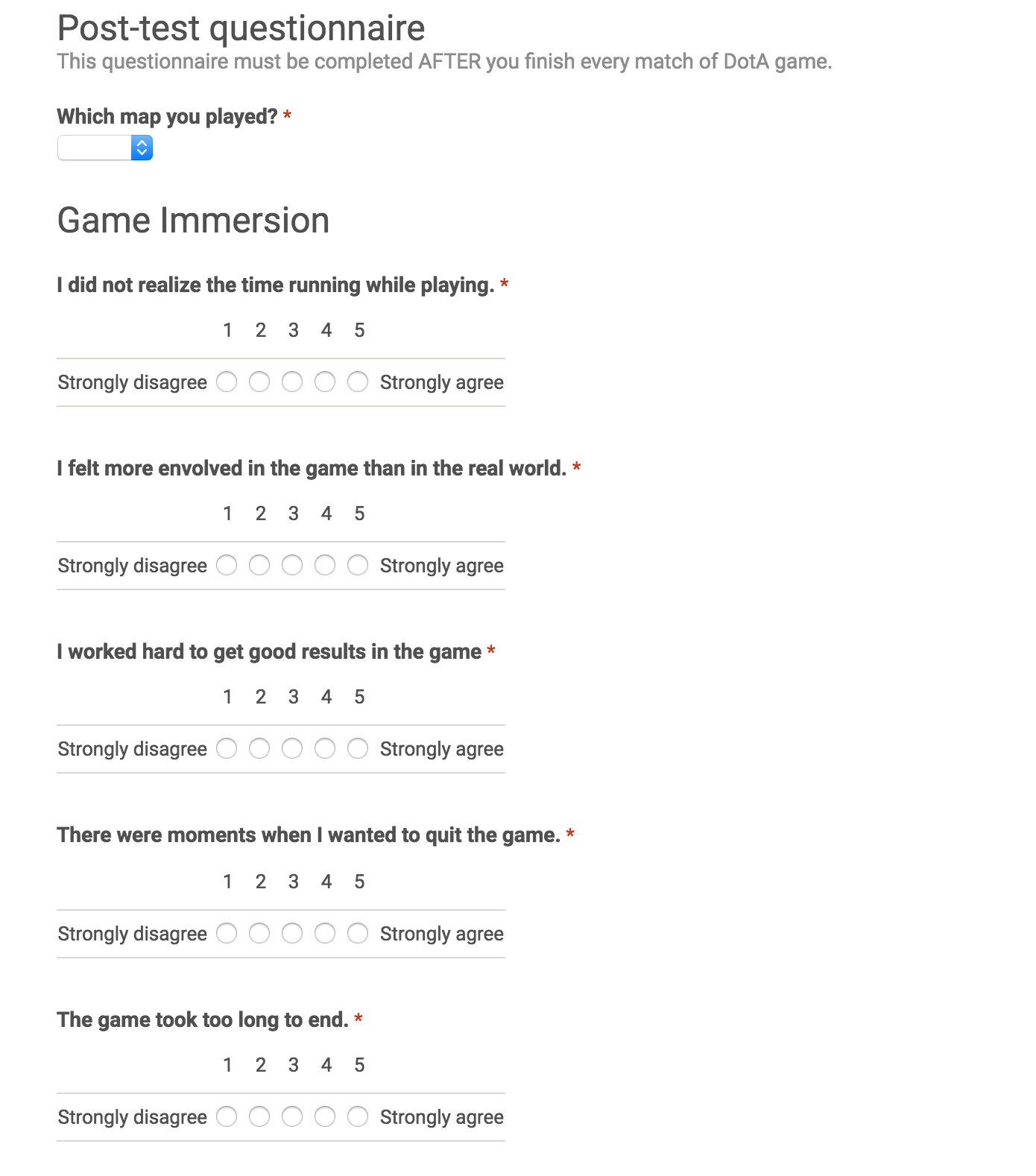}}
\caption{Post-test questionnaire: Questions about immersion.}
\label{fig:form_pre2}
\end{figure}

\begin{figure}[h]
\centering
\fbox{\includegraphics[width=13cm]{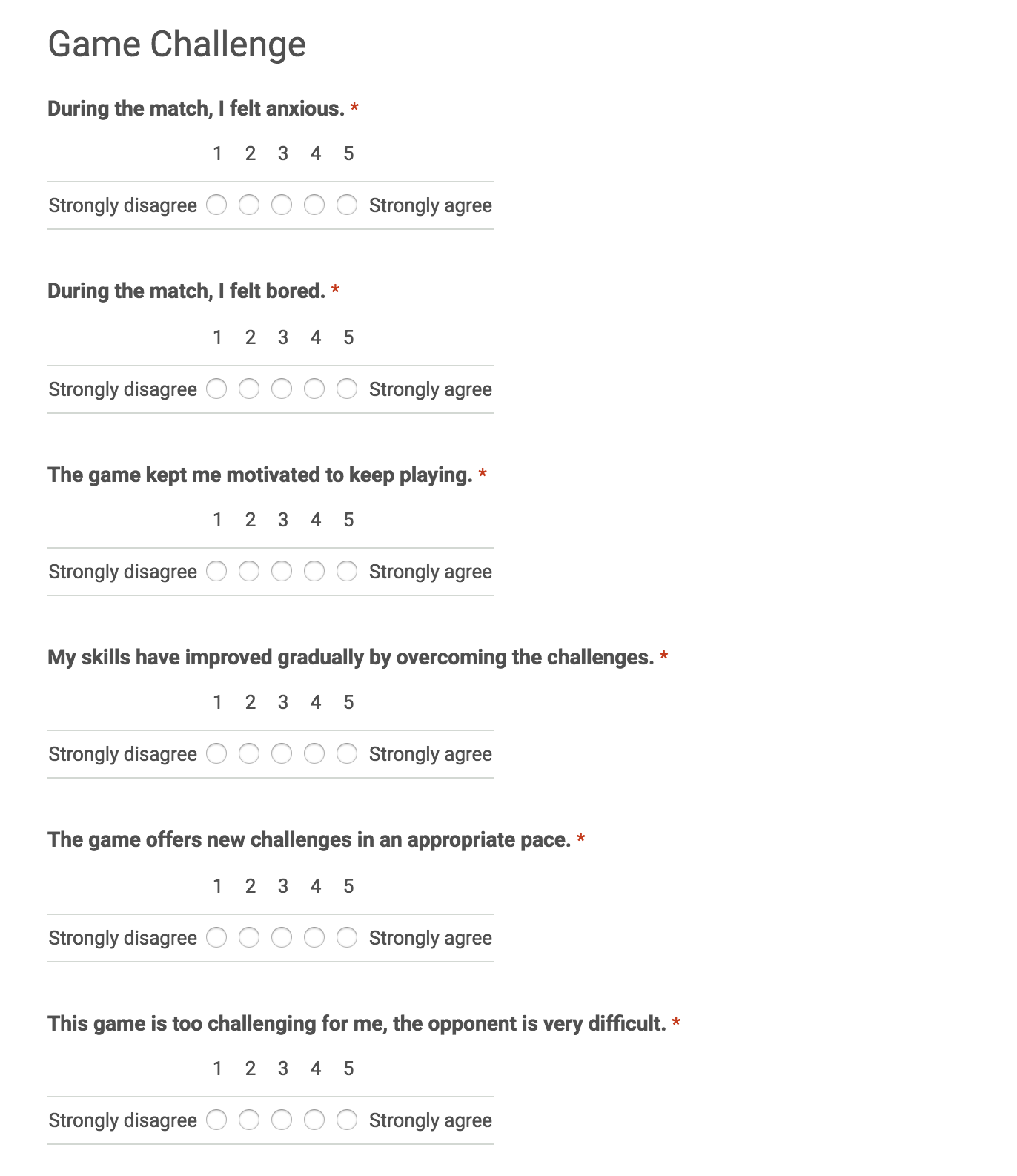}}
\caption{Post-test questionnaire: Questions about challenge.}
\label{fig:form_pos1}
\end{figure}

\begin{figure}[h]
\centering
\fbox{\includegraphics[width=13cm]{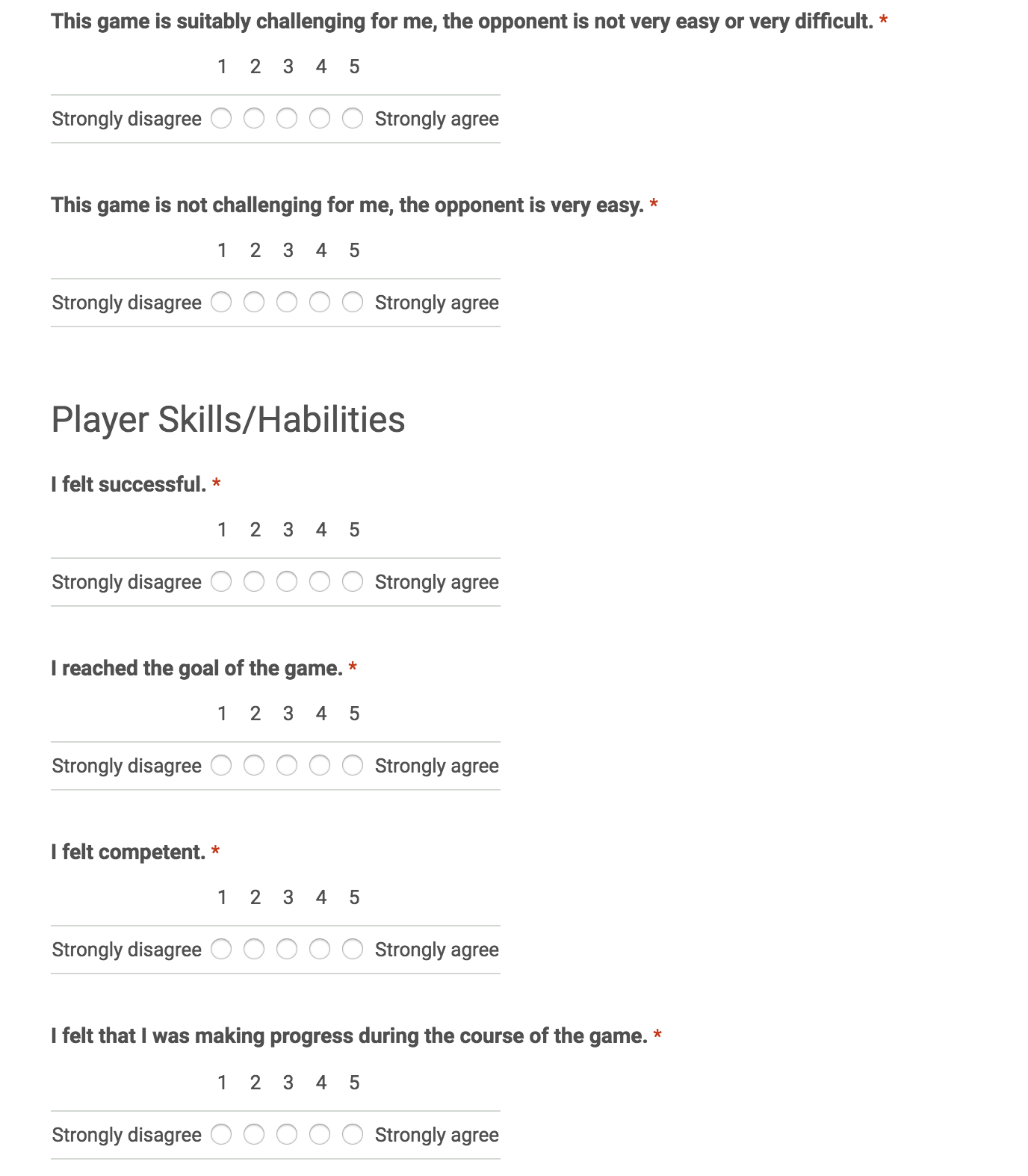}}
\caption{Post-test questionnaire: Questions about player skills.}
\label{fig:form_pos2}
\end{figure}

\begin{figure}[h]
\centering
\fbox{\includegraphics[width=15cm]{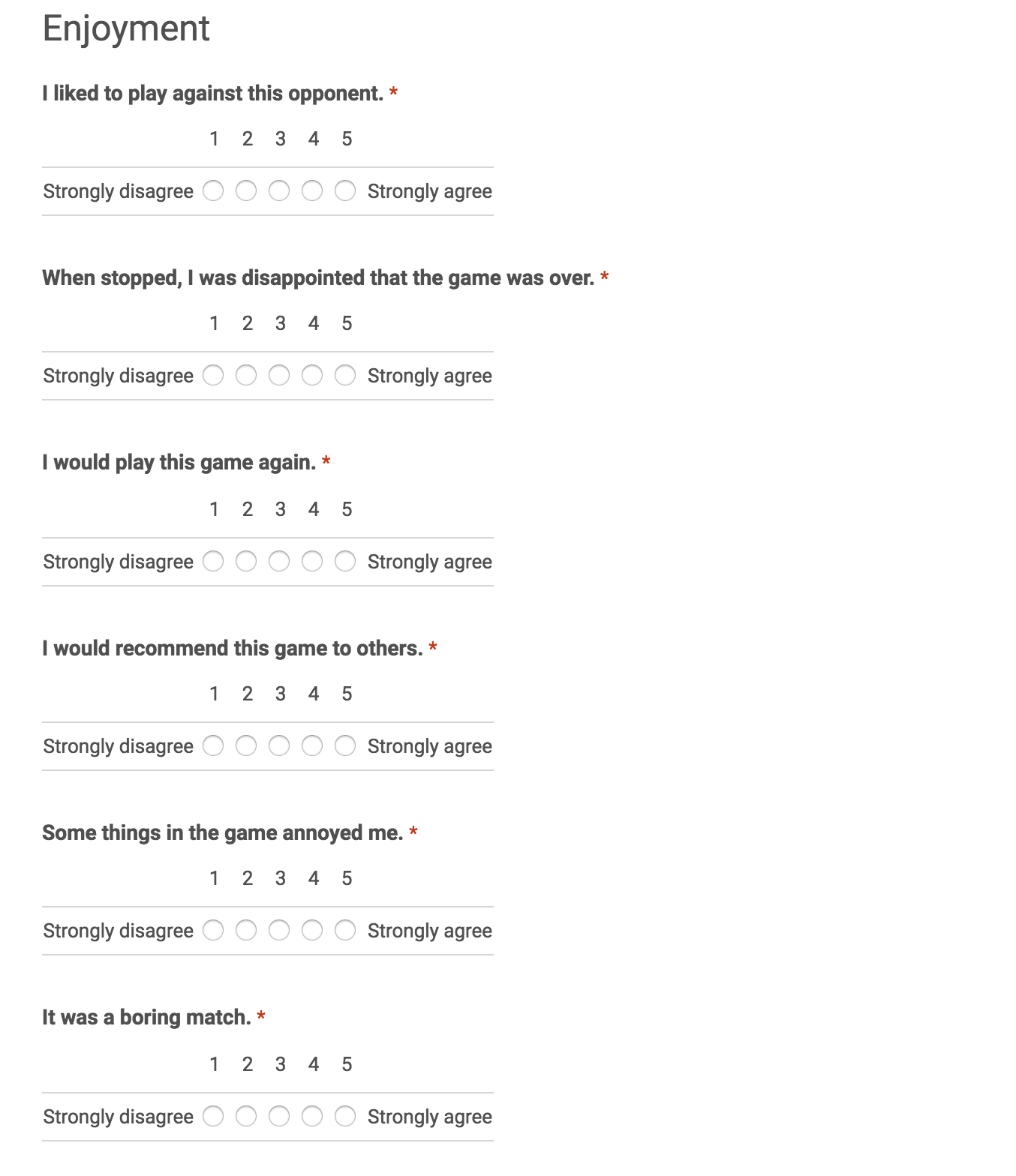}}
\caption{Post-test questionnaire: Questions about enjoyment during the match.}
\label{fig:form_pos3}
\end{figure}

\begin{figure}[h]
\centering
\fbox{\includegraphics[width=13cm]{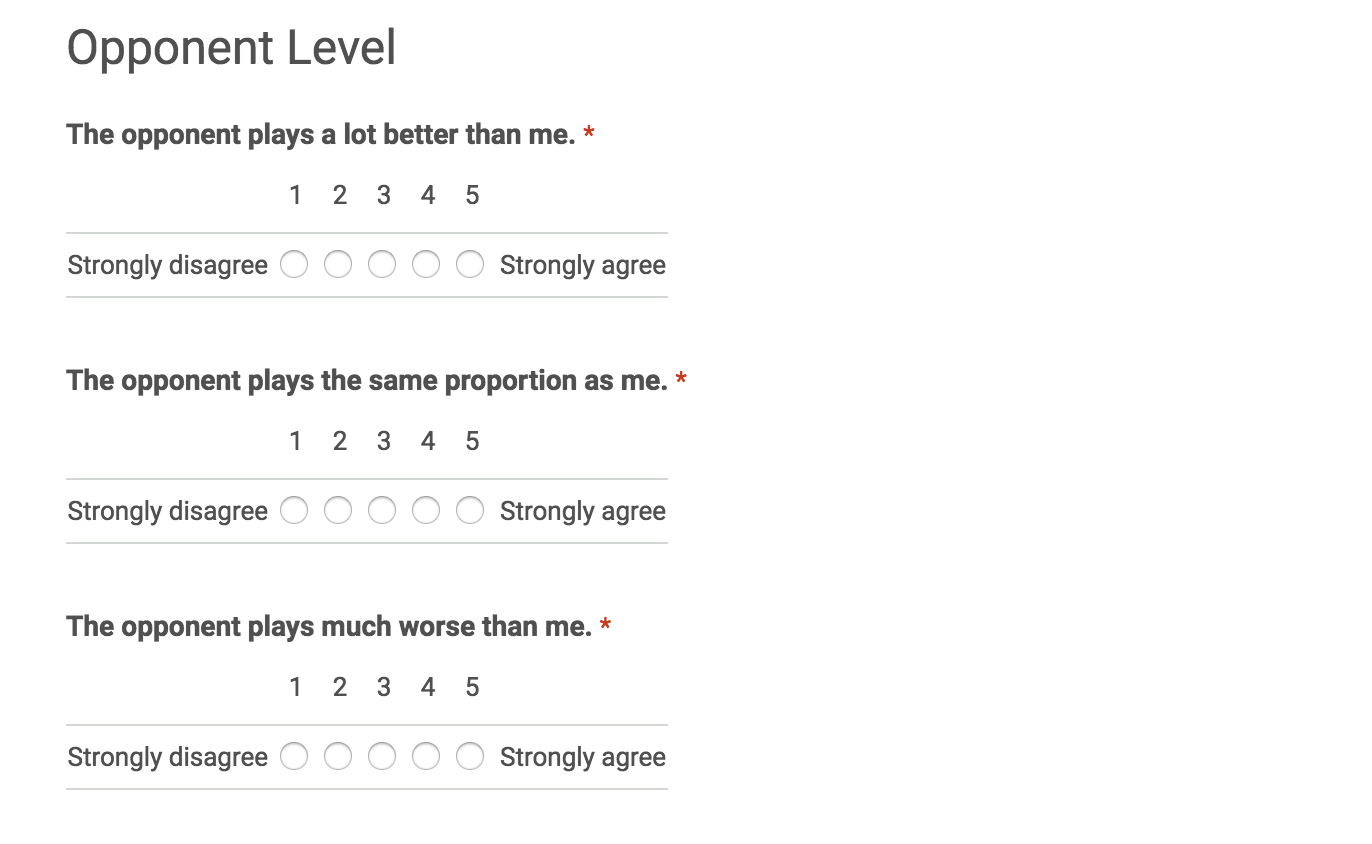}}
\caption{Post-test questionnaire: Questions about the opponent level.}
\label{fig:form_pos4}
\end{figure}

\begin{figure}[h]
\centering
\fbox{\includegraphics[width=13cm]{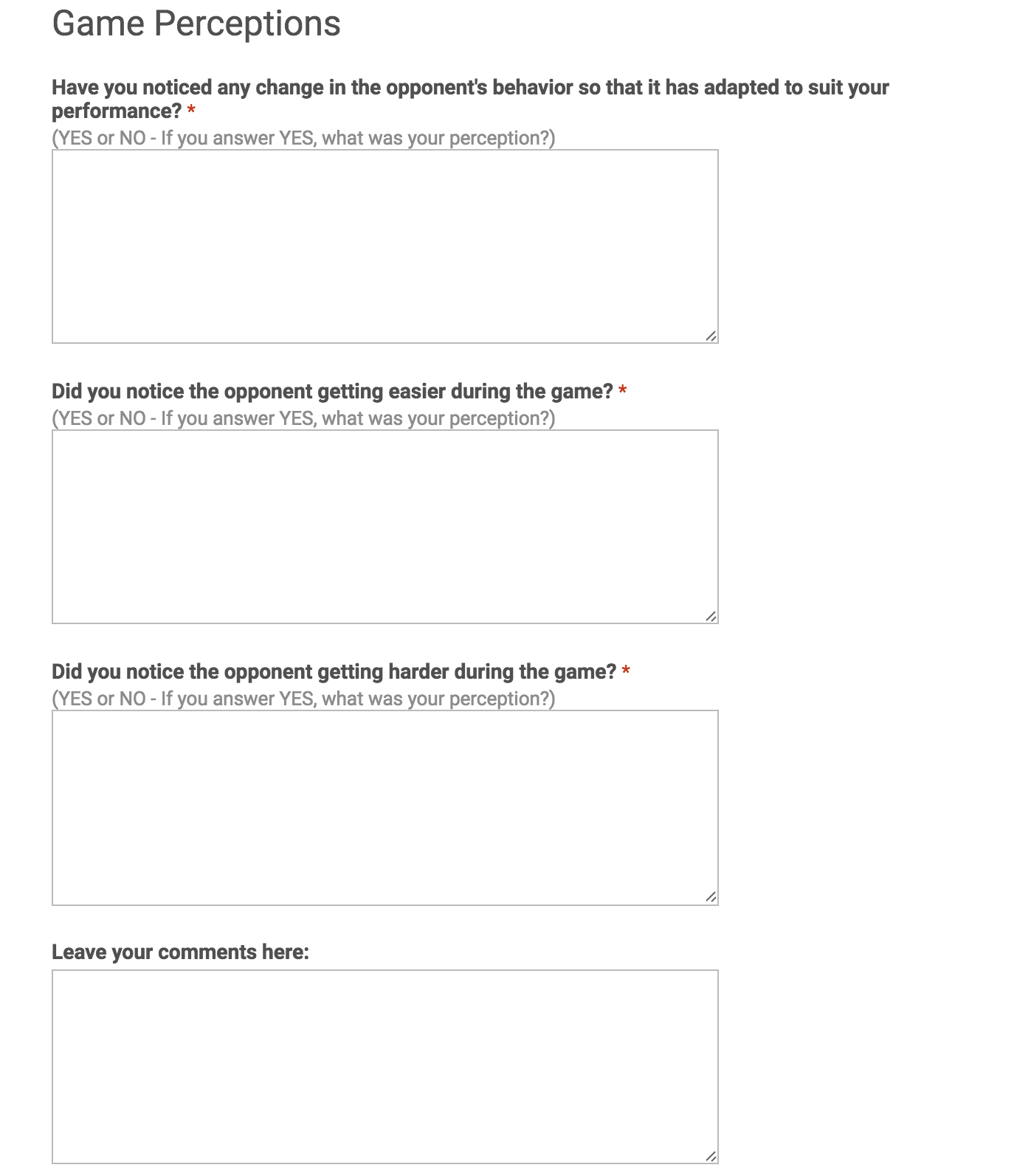}}
\caption{Post-test questionnaire: Questions about the player's perception during the game match.}
\label{fig:form_pos5}
\end{figure}


\end{document}